\documentclass{article}

\usepackage{microtype}
\usepackage{graphicx}
\usepackage{subfigure}
\usepackage{booktabs} 
\usepackage{multirow}
\usepackage{cite}
\usepackage{makecell}
\usepackage{epsfig,amsmath,amssymb,amsthm,array}
\usepackage[round,longnamesfirst]{natbib}

\newtheorem{theorem}{Theorem}

\usepackage{algorithmic}
\usepackage{algorithm}
\usepackage{hyperref}
\usepackage{url}

\usepackage[compact]{titlesec}
\titlespacing{\section}{0pt}{2ex}{1ex}
\titlespacing{\subsection}{0pt}{1ex}{0ex}
\titlespacing{\subsubsection}{0pt}{0.5ex}{0ex}
\usepackage[accepted]{icml2021}

\icmltitlerunning{Noise Optimization for Artificial Neural Networks}
\begin{document}
\twocolumn[
\icmltitle{Noise Optimization for Artificial Neural Networks}



\icmlsetsymbol{equal}{*}

\begin{icmlauthorlist}
\icmlauthor{Li Xiao}{equal,ict}
\icmlauthor{Zeliang Zhang}{equal,pku,hust}
\icmlauthor{Yijie Peng}{pku}

\end{icmlauthorlist}

\icmlaffiliation{ict}{Key Laboratory of Intelligent Information Processing, Institute of Computing Technology, Chinese Academy of Sciences, Beijing, 100090 China}
\icmlaffiliation{pku}{Department of Management Science and Information Systems, Guanghua School of Management, Peking University, Beijing,100871 China}
\icmlaffiliation{hust}{School of computer science and technology, Huazhong University of Science and Technology}

\icmlcorrespondingauthor{Yijie Peng}{pengyijie@pku.edu.cn}
\icmlcorrespondingauthor{Li Xiao}{andrew.lxiao@gmail.com, xiaoli@ict.ac.cn}
\icmlkeywords{Machine Learning, ICML}

\vskip 0.3in
]



\printAffiliationsAndNotice{\icmlEqualContribution} 

\begin{abstract}
Adding noises to artificial neural network(ANN) has been shown to be able to  improve robustness in previous work. In this work, we propose a new technique to compute the pathwise stochastic gradient estimate with respect to the standard deviation of the Gaussian noise added to each neuron of the ANN. By our  proposed technique, the gradient estimate with respect to noise levels is a byproduct of the backpropagation algorithm for estimating gradient with respect to synaptic weights in ANN. Thus, the noise level for each neuron can be optimized simultaneously in the processing of training the synaptic weights at nearly no extra computational cost. In numerical experiments, our proposed method can achieve significant performance improvement on robustness of several popular ANN structures under both black box and white box attacks tested in various computer vision datasets. code is available at \url{https://github.com/LX-doctorAI/noiseoptimize}
 
\end{abstract}

\section{Introduction}
Artificial neural network(ANN)s have been widely used in image processing, speech recognition, game, and medical diagnosis. However, ANNs are typically  vulnerable to adversarial attacks ~\cite{szegedy2013intriguing}. Many previous papers propose to add noises into ANN for improving robustness \cite{neelakantan2015adding,gulcehre2016noisy,brownlee2019train,you2019adversarial}. Adding noises to ANN may flatten the local minima and thus leads to robustness enhancement.

Adversarial attacks are small perturbations generated by computer algorithms. The small perturbations added to the input data can drastically alter the output of ANN ~\cite{REN2020}, which poses a serious challenge in security-critical applications, such as face recognition~\cite{parkhi2015deep}  and autonomous driving~\cite{hadash2018estimate}. 
On the other hand, human vision system is surprisingly robust under rather subtle structural changes, let alone the small computer-generated perturbations ~\cite{azulay2019why}, natural noise corruptions such as snow, blur, pixelation, and even their combinations. Therefore, achieving the human-like robustness is still a holy grail in computer vision research. 

There are evidences showing that proper regularization methods can effectively improve robustness of ANN under adversarial attacks. Previous work ~\cite{krizhevsky2012imagenet,you2019adversarial} adds noises to ANN for improving robustness, which can be viewed as a regularization method to alleviate over-fitting. However, the magnitudes of the injected noises are set in an ad-hoc manner in previous work. Our work is aligned with the previous work in terms of adding noises for improving robustness. 
The main methodological contribution of our work lies in proposing a new technique to compute the pathwise stochastic gradient estimate with respect to the standard deviation of the Gaussian noise added to each neuron of the ANN. By our proposed technique, the gradient estimate with respect to noise levels is a byproduct of the backpropagation (BP) algorithm for estimating gradient with respect to synaptic weights in ANN. Thus, the noise level for each neuron can be optimized simultaneously in the processing of training the synaptic weights at nearly no extra computational cost.

The pathwise stochastic gradient estimation technique is also known as infinitesimal perturbation analysis (IPA) in simulation literature (\citealp{asmussen2007stochastic}). IPA and the likelihood ratio (LR) method are two classic unbiased stochastic gradient estimation techniques  (\citealp{ho1991discrete}, \citealp{rubinstein1993discrete}). Recent advances can be found in \citet{hong2009estimating}, \citet{heidergott2010weak}, and \citet{peng2015discontinuity}. Stochastic gradient estimation has been a central topic in simulation optimization, and recently, a comprehensive review paper is written by a research team of Google's DeepMind \cite{mohamed2019monte}.

The proposed new method is implemented to train multi-layer perceptron (MLP) and convolution neural network (CNN) with a ResNet backbone in MNIST, Cifar-10 and tiny-ImageNet datasets. We test the performance under both white box and  and black box attacks. For black box attacks, we add both adversarial attacks and natural noise corruptions to the images. All numerical experiments show that our method can significantly improve robustness of ANN in nearly all situations, and it also improves classification accuracy in original dataset.

\section{Related work}
Adversarial attacks can be categorized to three types, i.e, 1) black box attack~\cite{hang2020ensemble,papernot2016distillation,guo2019simple}, where the attacker has no information about the internal structure of the attacked model, training parameters, and defense methods (if defense methods are used), and it can only interact with the model through outputs; 2) white box attack~\cite{dong2018boosting,nazemi2019potential}, where the attacker has full information about the attacked model; 3) gray box attack~\cite{prabhu2018grey,xiang2020side}, where the attacker only has a partial information of the model. 

Researchers have developed many gradient-based adversarial attack methods, such as L-BFGS~\cite{szegedy2013intriguing}, FGSM~\cite{goodfellow2014explaining}, and PGD ~\cite{madry2017towards}. The PGD attack is the strongest first-order attack that utilizes local information of the ANN. These methods are white box attacks in their original designs, but  they can also work as gray box attacks and black box attacks due to the transferability of adversarial attacks among models~\cite{Florian207,petrov2019measuring}. 

Compared with adversarial samples, adding natural noises to corrupt the input is a simpler black box attack~\cite{heaven2019deep,borji2019white}. Various types of  natural noises such as Gaussian, Impulse, Contrast, Elastic, and Blurs  have been developed  ~\cite{vasiljevic2016examining,zheng2016improving,Hendrycks2019noise}. \citet{Hendrycks2019noise} proposed a new metric to evaluate robustness under several types of natural noise corruptions. Each type of noise has $5$ severity levels and the evaluation metric is the average accuracy under noise corruptions  at $5$ severity levels. 

Previous work focusing on improving robustness under adversarial attacks includes feature squeezing~\cite{xu2017feature}, distillation network~\cite{papernot2016distillation}, input transformation (e.g., JPEG compression ~\cite{jpeg} , autoencoder-based denoising~\cite{liao2018defense}  and regularization~\cite{ross2017improving} ), Parseval network~\cite{cisse2017parseval}, gradient  masking~\cite{masking},  randomization~\cite{liu2018towards,dhillon2018stochastic},  radial  basis  mapping  kernel~\cite{taghanaki2019kernelized},  non-local  context  encoder~\cite{he2019non}, and Per~\cite{dong2019benchmarking}. The PGD-based adversarial (re)training, which  augments the training  set  with  adversarial  examples,  is  the  most effective  defense  strategy~\cite{goodfellow2014explaining,tramer2017ensemble,PGD},  but it consumes too much training time and can be neutered  completely or partially by adaptive attacks~\cite{CW_bpda,CW_ten,tramer2020adaptive}.

Other previous work improves robustness via adding noises to input data ~\cite{hendrycks2019augmix, gao2020fuzz}, or adding noises to activations, outputs, weights and even gradients \cite{neelakantan2015adding,gulcehre2016noisy,brownlee2019train,you2019adversarial,xiao2019training}. None of the previous work considers how to optimally set the magnitudes of the noises added to the ANN. 

\section{Noise Optimization Method}
\subsection{Gradient Estimation}
Let $\tau$ denote the number of layers in the neural network and $m_{t}$ denote the number of neurons in the $t$-th layer, $t\in{1, 2,..., \tau}$. We denote the output of $t$-th layer as $\bf{X}^{(t)}=[\bf{x}^{(t)}_{1}, \bf{x}^{(t)}_{2}, ..., \bf{x}^{(t)}_{m_{t}}] \in \mathbb{R}^{m_{t}}$ and $\bf{X}^{(0)}$ is the input of the network.  

\begin{figure*}[h]
	\centering
	\includegraphics[scale=0.055]{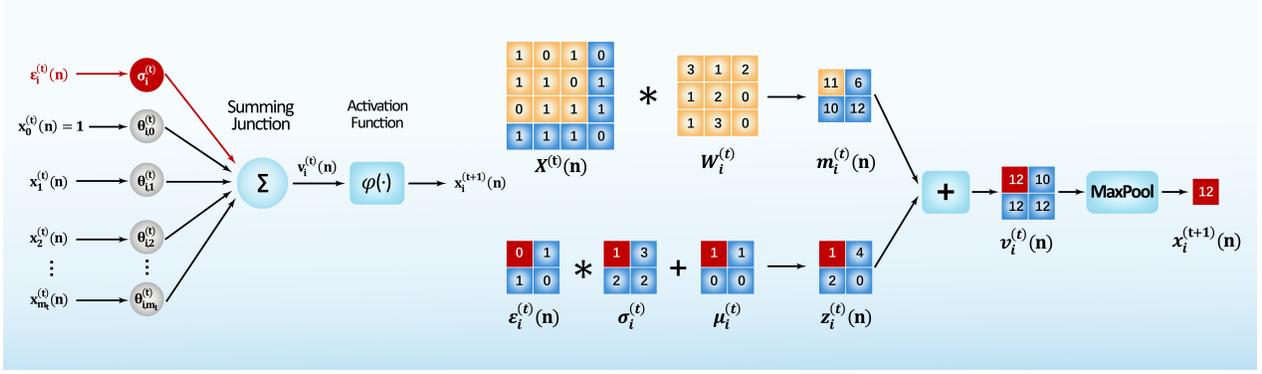}
	\label{forward_example}
	\caption{Illustration for the forward propagation of ANN with noises. The left-hand side of the figure shows computation in MLP, and the right-hand side of the figure shows computation in CNNs.}
\end{figure*}

Suppose we have $N$ inputs for the network, denoted as $\bf{X}^{(0)}(n)$, $n=1, 2, ..., N$. For the $n$-th input, the $i$-th output at the $t$-th layer can be given by

\begin{equation}\label{fw}
    \bf{x}^{(t+1)}_{i}(n) = \varphi(\bf{v}^{(t)}_{i}), ~~~ \bf{v}^{(t)}_{i}=\sum_{j=0}^{m_{t}}\theta^{(t)}_{i,j}\bf{x}^{(t)}_{j}(n)+\bf{z}^{(t)}_{i}(n),
\end{equation}
where $\bf{x}^{(t)}_{j}(n)$ is the $j$-th input at the  $t$-th layer for the $n$-th data, $\theta^{(t)}_{i,j}$ is the weight for the $i$-th input at the $t$-th layer,  $\bf{v}^{(t)}_{i}$ is the $i$-th logit output at the $t$-th layer, $\varphi$ is the activation function and $\bf{z}^{(t)}_{i}(n)$ is an independent random noise added to the $i$-th neuron at the $t$-th layer for the $n$-th data. The computation of Eq.(\ref{fw}) is depicted on the right-hand side of Figure \ref{forward_example}. 
 We let $\bf{x}_0^{(t)}(n)\equiv 1$ and then $\theta^{(t)}_{i,0}$ is the bias term in the linear operation of the $i$-th neuron at the $t$-th layer.  

Computation in CNN, which is depicted on the right-hand side of Figure \ref{forward_example}, is essentially equivalent to the computation  of Eq.(\ref{fw}) in MLP. In Figure \ref{forward_example}, the orange colored element in $\bf{m}^{(t)}_{i}$ of the $i$-th feature map at the $t$-th layer is a product of the parameters $W_i^{(t)}$ in the $i$-th convolution kernel and the corresponding orange colored inputs in $\bf{X}^{(t)}(n)$. This computation is equivalent to the linear operation on the inputs of a neuron in Eq.(\ref{fw}). An independent normal random variable is added to each element in the feature map. Mean $\mu_i^{(t)}$ can be viewed as the bias term in Eq.(\ref{fw}). 

\begin{figure*}[t]
	\centering
	\includegraphics[height=2in, width=6.5in]{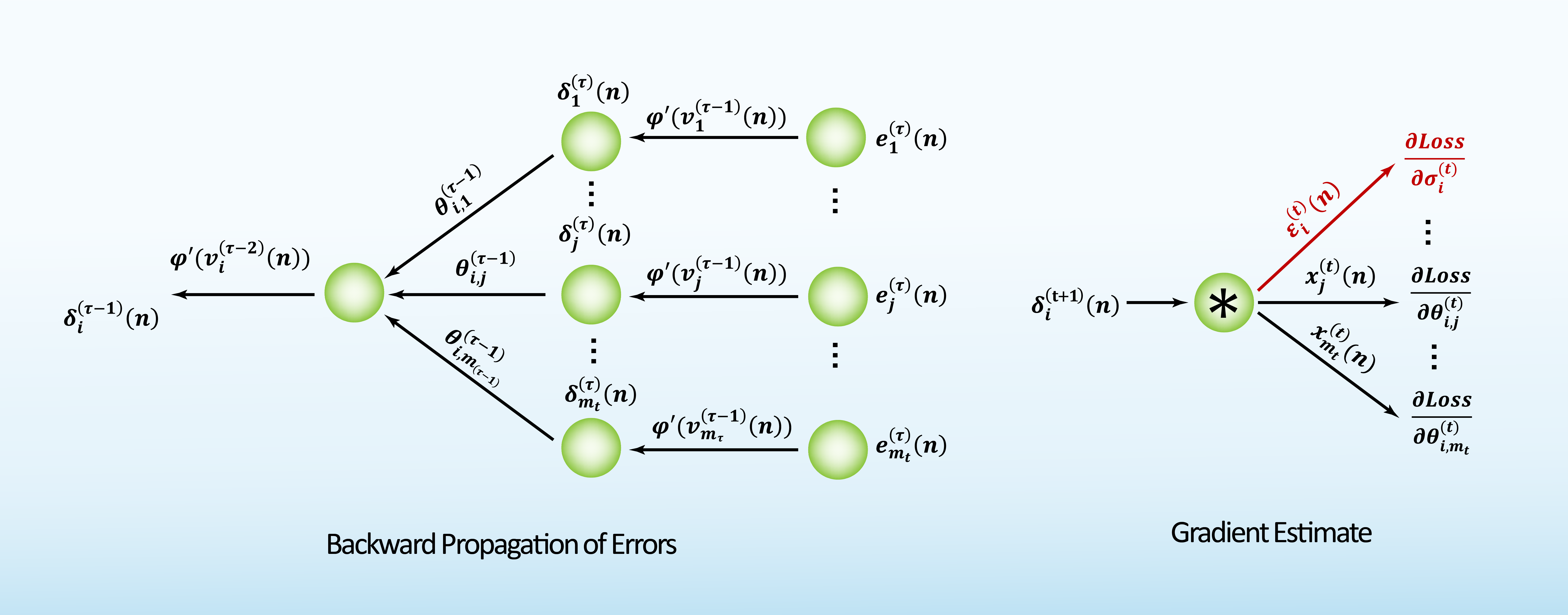}
	\label{backward_example}
	\caption{ The left-hand side of the figure presents the backward propagation of residual errors, and the right-hand side the figure shows computation of the gradient estimate based on residual errors. }
\end{figure*}

We denote the loss function as $\mathcal{L}$. For the $n$-th data $\bf{X}^{(0)}(n)$ with label $\bf{Y}(n)$, we have the loss value denoted by $\mathcal{L}(\bf{X}^{(\tau)}(n), \bf{Y}(n))$.
In our work, we try to optimize the magnitude of the noise level $\sigma^{(t)}_{i}$ for centered normal random noise added to each neuron, i.e., $\bf{z}^{(t)}_{i}(n)=\sigma^{(t)}_{i}\varepsilon^{(t)}_{i}(n)$, where $\varepsilon^{(t)}_{i}(n)$ is a standard normal random variable. 
Define the residual error for the $i$-th neuron at the $t$-th layer  for the $n$-th data propagated backwardly through the ANN as
	\begin{equation}\label{bp}
\begin{aligned}
\bf{\delta}_i^{(t)}(n)=\begin{cases}
\bf{e}_{i}^{(\tau)}(n)\varphi'\left(\bf{v}_i^{(\tau-1)}(n)\right),\qquad t=\tau,
\\
\varphi'\left(\bf{v}_i^{(t-1)}(n)\right)\left( \sum_{j=0}^{m_{t}}\theta_{j,i}^{(t)} \bf{\delta}_j^{(t+1)}(n)\right),~t<\tau,
\end{cases}
\end{aligned}
\end{equation}
where $\bf{e}_{i}^{(\tau)}(n)$ is defined by 
\begin{align*}
\bf{e}_{i}^{(\tau)}(n):=\left.\frac{\partial \mathcal{L}(\bf{x},\bf{Y}(n))}{\partial \bf{x}_{i}}\right|_{x=\bf{X}^{(\tau)}(n)}~.
\end{align*}
The computation of residual errors by BP is depicted on the left-hand side of Figure \ref{backward_example}. 
 The BP algorithm essentially offers  pathwise stochastic derivative estimates for the loss $\mathcal{L}$ with respect to all parameters $\theta^{(t)}_{i,j}$, $t=1,2,\ldots,\tau-1$, $j\in\{0,1,\ldots,m_t \}$, $i\in\{0,1,\ldots,m_{t+1}\}$  simultaneously. Specifically,  
 	\begin{align*}
 \frac{\partial \mathcal{L}(\bf{X}^{(\tau)}(n),\bf{Y}(n))}{\partial \theta_{i,j}^{(t)}}=\bf{\delta}_{i}^{(t+1)}(n) \bf{x}_{j}^{(t)}(n).
 \end{align*}

 In the following Theorem 1, we show that the pathwise stochastic derivatives with respect to the magnitudes of the noise levels $\sigma^{(t)}_{i}$, $i=1\ldots,m_t$, $t=1,\ldots,\tau$, can be estimated as a byproduct of the BP algorithm, and they can be computed in a similar matter as the the pathwise stochastic derivatives with respect to synaptic weights. The computation of pathwise stochastic derivatives  is depicted on the right-hand side of Figure \ref{backward_example}.

\begin{theorem} 
	Assume the activation function $\varphi$ and the loss function $\mathcal{L}$ are differentiable. We have
	\begin{equation}\label{pd}
	\begin{aligned}
	&\frac{\partial \mathcal{L}(\bf{X}^{(\tau)}(n), \bf{Y}(n))}{\partial \sigma^{(t)}_{i}}\\
	&=\frac{1}{\sigma^{(t)}_{i}}\bf{\delta}_{i}^{(t+1)}(n) \bf{z}_{i}^{(t)}(n)=\bf{\delta}_{i}^{(t+1)}(n)\bf{\epsilon}_{i}^{(t)}(n).
	\end{aligned}
	\end{equation}
\end{theorem}

\proof
 The pathwise stochastic derivative  for the sensitivity with respect to $\sigma^{(t)}_{i}$ is 
\begin{equation}
\small
   \sum\limits^{m_{\tau}}_{j=0}\frac{\partial{\mathcal{L}(\bf{x},\bf{Y}(n))}}{\partial{\bf{x}_{j}}}\big{\vert}_{\bf{x}=\bf{X}^{(\tau)}(n)}\frac{\bf{x}^{(\tau)}_{j}(n)}{\partial\sigma^{(t)}_{i}},
 \label{base_p}
\end{equation}
where  
\begin{align*}
\small
&\frac{\partial\bf{x}^{(l+1)}_{j}(n)}{\partial\bf{\sigma^{(t)}_{i}}}=\varphi'(\bf{v}^{(l)}_{j}(n))\frac{\partial\bf{v}^{(l)}_{j}(n)}{\partial\sigma^{(t)}_{i}},\\
&\frac{\partial\bf{v}^{(l)}_{j}(n)}{\partial\bf{\sigma^{(t)}_{i}}}=\sum\limits^{m_{t}}_{\ell=1}\bf{\theta}^{(t)}_{j,\ell}\frac{\partial\bf{x}^{(l)}_{\ell}(n)}{\partial\sigma^{(t)}_{i}}+\frac{\partial\sigma^{(l)}_{j}}{\partial\sigma^{(t)}_{i}}\epsilon^{(t)}_{j}(n). 
\end{align*}

Notice that 
\begin{align*}
\small
    &\frac{\partial\bf{v}^{(l)}_{j}(n)}{\partial\bf{\sigma^{(t)}_{i}}}=\epsilon^{(t)}_{i}(n), ~~l=t,~ j=i,\\
     &\frac{\partial\bf{v}^{(l)}_{j}(n)}{\partial\bf{\sigma^{(t)}_{i}}}=\sum\limits^{m_{l}}_{\ell=1}\theta^{(l)}_{j,\ell}\frac{\partial\bf{x}^{(l)}_{\ell}(n)}{\partial\sigma^{(t)}_{i}}, j \neq i ~\mbox{or}~ l \neq t.
\end{align*}

Then the pathwise stochastic derivative on the left-hand side of Eq.(\ref{base_p}) can be written as the following nested summations:
\begin{equation*}
\begin{aligned}
\small
    &\sum\limits^{m_{\tau}}_{i_{\tau}=1}\bf{e}^{(\tau)}_{i_{\tau}}(n)\varphi'(\bf{v}^{(\tau-1)}_{i_{\tau}}(n))\\
&\times \{\sum\limits^{m_{\tau-1}}_{i_{\tau-1}=1}\theta^{(\tau-1)}_{i_{\tau},i_{\tau-1}}[\times ... \times[\sum\limits^{m_{t+1}}_{i_{t+1}=1}\theta^{(t+1)}_{i_{t+2},i_{t+1}}\varphi'(\bf{v}^{(t+1)}_{i_{t+1}}(n))\\
&\times(\sum\limits^{m_{t}}_{i_{t}=1}\theta^{(t)}_{i_{t+1},i_{t}}\varphi'(\bf{v}^{(t)}_{i_{t}}(n))\epsilon^{(t)}_{i_{t}}(n))]]\}.
\end{aligned}
\end{equation*}

By reserving the order of summations, we obtain 
\begin{equation}
\small
    \begin{aligned}
        &\sum\limits^{m_{t}}_{i_{t}=1}\epsilon^{(t)}_{i_{t}}(n)\varphi'(\bf{v}^{(t)}_{i_{t}}(n))\theta^{(t)}_{i_{t+1,i_{t}}}\times\{\sum\limits^{m_{t+1}}_{i_{t+1}=1}\theta^{(t+1)}_{i_{t+2,i_{t+1}}}\\&\times\varphi'(\bf{v}^{(t+1)}_{i_{t+1}}(n))[\times...\sum\limits^{m_{\tau}}_{i_{\tau}=1}\theta^{(\tau-1)}_{i_{\tau},i_{\tau-1}}\bf{e}^{(\tau)}_{i_{\tau}}(n)\varphi'(\bf{v}^{(\tau-1)}_{i_{\tau}}(n))]\}        
    \end{aligned}
\end{equation}
which leads to the right-hand side of Eq.(\ref{pd}) by the definition of  residual error $\bf{\delta}_i^{(t)}(n)$. 
\endproof

Next we show that the pathwise stochastic derivative  Eq.(\ref{pd}) is an  unbiased estimate for the derivative of the expected loss over the randomness in ANN. The key is to justify the interchange of derivative and expectation. 

\begin{theorem} Assume $\varphi(\cdot)$ and $\mathcal{L}$ are differentiable almost everywhere, and \begin{align}\label{c1}
	\mathbb{E}\left[\sup_{\bf{\sigma}_{i}^{(t)}\in\bf{\Sigma}_{i}^{(t)}}|\bf{\delta}_i^{(t+1)}(n)|\right]<\infty,
		\end{align}
	  where $\bf{\Sigma}_{i}^{(t)}$ is a neighborhood surrounding $\bf{\sigma}_{i}^{(t)}$. Then, 
	$$\frac{\partial}{\partial \bf{\sigma}^{(t)}_{i}}\mathbb{E}\left[ \mathcal{L}(\bf{X}^{(\tau)}(n), \bf{Y}(n)) \right]=\mathbb{E}\left[\frac{\partial \mathcal{L}(\bf{X}^{(\tau)}(n), \bf{Y}(n))}{\partial \bf{\sigma}^{(t)}_{i}}\right].$$
\end{theorem}
\proof For simplicity, we suppress the dependency on $n$ for notations in the proof. By definition, 
\begin{align*}
\small
&\frac{\partial}{\partial \bf{\sigma}^{(t)}_{i}}\mathbb{E}\left[ \mathcal{L}(\bf{X}^{(\tau)}[\bf{\sigma}^{(t)}_{i}], \bf{Y}) \right]\\
=&\lim_{\varepsilon\to 0} \frac{1}{\varepsilon}\left( \mathbb{E}\left[ \mathcal{L}(\bf{X}^{(\tau)}[\bf{\sigma}^{(t)}_{i}+\varepsilon], \bf{Y}) \right] -\mathbb{E}\left[ \mathcal{L}(\bf{X}^{(\tau)}[\bf{\sigma}^{(t)}_{i}], \bf{Y}) \right] \right)\\
=& \lim_{\varepsilon\to 0} \mathbb{E}\left[ \bf{\delta}_{i}^{(t+1)}[\bf{\sigma}^{(t)}_{i}+\lambda\varepsilon]\bf{\epsilon}_{i}^{(t)}\right]= \mathbb{E}\left[ \bf{\delta}_{i}^{(t+1)}[\bf{\sigma}^{(t)}_{i}]\bf{\epsilon}_{i}^{(t)}\right]\\
=&\mathbb{E}\left[\frac{\partial \mathcal{L}(\bf{X}^{(\tau)}[\bf{\sigma}^{(t)}_{i}], \bf{Y})}{\partial \bf{\sigma}^{(t)}_{i}}\right],
\end{align*}
where $Z[\cdot]$ denotes a quantity $Z$ dependent on the argument, the second equality holds by applying the mean-value theory to the conclusion of Theorem 1 with $\lambda\in(0,1)$, and the third equality holds due to the dominated convergence theorem to justify the interchange of limit and expectation under uniform integrability condition Eq.(\ref{c1}) on the residual error. 
\endproof

\subsection{Gradient-Based Searching Method}
To reduce oscillation in the gradient-based search, we apply the Adam optimizer to update $\sigma^{(t)}_{i}$  as follows
\begin{equation}
\small
\begin{aligned}
    \sigma^{(t)}_{i} &\gets \vert ~\sigma^{(t)}_{i} - \alpha * \frac{\hat{h}^{(t)}_{i}}{\sqrt{\hat{v}^{(t)}_{i}}+\varepsilon}~\vert,\\
     \hat{h}^{(t)}_{i} &\gets \frac{h^{(t)}_{i}}{1-\beta^{N}_{1}},\\
     h^{(t)}_{i} &\gets \beta_{1} * h^{(t)}_{i} + (1-\beta_{1}) * \bf{g}^{(t)}_{i}(n),\\
    \hat{v}^{(t)}_{i} &\gets \frac{v^{(t)}_{i}}{1-\beta^{N}_{2}},\\
    v^{(t)}_{i} &\gets \beta_{2} * v^{(t)}_{i} + (1-\beta_{2}) * \bf{g}^{(t)}_{i}(n) * \bf{g}^{(t)}_{i}(n),
\end{aligned}
\label{noise_optimization_equation}
\end{equation}
where $N$ is the current number of iterations, and the absolute value is taken in the update of $\sigma^{(t)}_{i}>0$ to enforce constraint $\sigma^{(t)}_{i}>0$. 
We set $\beta_{1} = 0.9$, $\beta_{2} = 0.999$, $\varepsilon = 10^{-8}$, and $m^{(t)}_{i}$ and $v^{(t)}_{i}$ are the exponential moving average of the gradient and its square, respectively, and $\bf{g}^{(t)}_{i}(n)$ is the pathwise stochastic derivative estimate with respect to $\sigma^{(t)}_{i}$ derived in the last subsection.  To avoid rapid change of $\sigma^{(t)}_{i}$ in  iterations, the initial learning rate is set at a relative low value $10^{-3}$.
The detailed training procedure is summarized in \textbf{Algorithm} \ref{alg:noise_optimized}.

\begin{algorithm}[tb]
   \caption{Noise Optimization for ANNs}
   \label{alg:noise_optimized}
\begin{algorithmic}[1]
   \STATE {\bfseries Input:} Training data  $D={\{(\mathbf{X}^{(0)}(n),\mathbf{Y(n)})\}}^{N}_{n=1}$, loss function $\mathcal{L}$.
   \STATE Construct a $\tau$-layers ANN and initialize all the parameters in ANNs.
   \REPEAT
   \STATE Exploiting Eq.~(\ref{fw}) to calculate the output $\mathbf{X}^{(\tau)}(n)$;
   \STATE Calculate loss function $\mathcal{L}(\mathbf{X}^{(\tau)}(n),\mathbf{Y}(n))$;
   \STATE Using Eq.~(\ref{bp}) and (\ref{pd}) to estimate the gradient of the loss respect to the weights and noise levels,  respectively;
   \STATE Updating the weights in ANN;
   \STATE Updating the noise level using Eq.~(\ref{noise_optimization_equation}).
   \UNTIL{Condition is met}
\end{algorithmic}
\end{algorithm}

\section{Experiments}

\begin{table*}[htb]
\centering
\small
\begin{tabular}{|l|c|c|c|c|c|c|}
\hline
Datasets &
  \multicolumn{1}{c|}{Models} &
  \multicolumn{1}{c|}{Optimizer} &
  \multicolumn{1}{c|}{Learning rate} &
  \multicolumn{1}{c|}{$L_{2}$ Weight decay} &
  \multicolumn{1}{c|}{Epochs} &
  \multicolumn{1}{c|}{Batch size} \\ \hline
\multirow{2}{*}{MNIST} &
  \multicolumn{1}{l|}{MLP} &
  \multirow{4}{*}{Adam} &
  \multirow{4}{*}{$1\times10^{-3}$} &
  \multirow{4}{*}{$1\times10^{-4}$} &
  \multirow{2}{*}{30} &
  \multirow{6}{*}{128} \\ \cline{2-2}
                          & \multicolumn{1}{l|}{CNN}  &                                   &                                                   &  &                     &  \\ \cline{1-2} \cline{6-6}
\multirow{2}{*}{Cifar-10} & \multirow{2}{*}{ResNet18} &                                   &                                                   &  & \multirow{2}{*}{50} &  \\
                          &                           &                                   &                                                   &  &                     &  \\ \cline{1-6}
\multirow{2}{*}{Tiny-ImageNet} &
  \multirow{2}{*}{ResNet34} &
  SGD &
  $1\times10^{-2}$ initially &
  \multirow{2}{*}{$5\times10^{-4}$} &
  \multirow{2}{*}{80} &
   \\ \cline{3-4}
                          &                           & \multicolumn{1}{l|}{momentum=0.9} & \multicolumn{1}{l|}{decay by 0.8 every 20 epochs} &  &                     &  \\ \hline
\end{tabular}
\caption{Summary of the hyper-parameters for training different models.}
\label{model_training}
\end{table*}

\begin{table*}[htb]

\centering
\begin{tabular}{|c|c|l|c|c|c|l|}
\hline
\multirow{2}{*}{Models} & \multicolumn{6}{c|}{Attack Methods}                                                  \\ \cline{2-7} 
                        & \multicolumn{2}{c|}{Corruption Noise} & FGSM   & L-BFGS   & \multicolumn{2}{c|}{PGD} \\ \hline
\multirow{2}{*}{MNIST} &
  \multicolumn{2}{c|}{\multirow{6}{*}{\begin{tabular}[c]{@{}c@{}}Including\\ Gaussian Noise,\\ Impulse Noise,\\  Glass Blur Noise, \\ and \\ Contrast Noise\end{tabular}}} &
  \multirow{4}{*}{$\alpha=0.1$} &
  \multirow{2}{*}{\begin{tabular}[c]{@{}c@{}}$\alpha=5\times10^{-1}$\\ N=10\end{tabular}} &
  \multicolumn{2}{c|}{\multirow{2}{*}{$\alpha=\frac{5}{255}$,$\epsilon=\frac{25}{255}$,N=10}} \\
                        & \multicolumn{2}{c|}{}                 &        &          & \multicolumn{2}{c|}{}    \\ \cline{1-1} \cline{5-7} 
\multirow{2}{*}{Cifar-10} &
  \multicolumn{2}{c|}{} &
   &
  \multirow{2}{*}{\begin{tabular}[c]{@{}c@{}}$\alpha=5\times10^{-2}$\\ N=20\end{tabular}} &
  \multicolumn{2}{c|}{\multirow{2}{*}{$\alpha=\frac{2}{255}$,$\epsilon=\frac{8}{255}$,N=5}} \\
                        & \multicolumn{2}{c|}{}                 &        &          & \multicolumn{2}{c|}{}    \\ \cline{1-1} \cline{4-7} 
\multirow{2}{*}{Tiny-ImageNet} &
  \multicolumn{2}{c|}{} &
  \multirow{2}{*}{$\alpha=\frac{2}{255}$} &
  \multirow{2}{*}{\begin{tabular}[c]{@{}c@{}}$\alpha=5\times10^{-2}$\\ N=10\end{tabular}} &
  \multicolumn{2}{c|}{\multirow{2}{*}{$\alpha=\frac{2}{255}$,$\epsilon=\frac{5}{255}$,N=3}} \\
                        & \multicolumn{2}{c|}{}                 &        &          & \multicolumn{2}{c|}{}    \\ \hline
\end{tabular}
\caption{Summary of the settings for different attack methods.}
\label{attack methods}
\end{table*}

\subsection{Datasets and experimental settings}

We conduct extensive experiments in three public datasets to test the robustness of our method: 1) the MNIST dataset; 2) the Cifar-10 dataset; 3) the tiny-Imagenet dataset (subset of the Imagenet)~\cite{le2015tiny}. For MNIST, we apply our method to  both Multi-Layer Perceptron (MLP) and convolution Neural networks(CNN). For Cifar-10, we use CNN with the ResNet18 backbone. For tiny-Imagenet, we use CNN with the ResNet34 backbone. All the codes are implemented based on the PyTorch 1.6.0 and run on Nvidia GeForce RTX 3090.

Both white box and black box attacks are used to test the robustness of ANN. For white box attacks, we apply FGSM~\cite{goodfellow2014explaining}, L-BFGS~\cite{szegedy2013intriguing} and PGD~\cite{madry2017towards} to generate adversarial samples. For black box attacks, we apply FGSM and L-BFGS with a different ANN structure than that under attack to generate adversarial samples. Unlike adversarial attacks which modify the pixels with the worst case perturbation, ~\citet{Hendrycks2019noise} propose to add various types of natural noises to the input images as corruption attacks. In our work, we also adopt four types of natural noises as black box attacks. For each type of noise, we compute the average accuracy of 5 strength levels of corruptions to evaluate  robustness.

The settings of hyper-parameters for training models, e.g. Optimizer, learning rate, weight decay, number of training  epochs and batch sizes are presented in Table \ref{model_training}. All setups are determined by a hyper-parameter search. The settings for the attack methods are presented in Table \ref{attack methods}. For FGSM, $\alpha$ is the step size. For L-BFGS, $\alpha$ is the step size and $N$ is the number of maximum iterations. For PGD, $\alpha$ is the step size, $\epsilon$ is the maximum permutaion in one pixel and $N$ is the number of maximum iterations. The settings of FGSM and PGD in MNIST and Cifar-10 datasets follow those set in previous work~\cite{ling2019deepsec,chan2020thinks}, whereas other setups are determined by a hyper-parameter search.

\subsection{Results in MNIST dataset}
We construct the MLP and CNN to train the MNIST dataset, respectively. The MLP contains two hidden layers with 100 and 50 neurons on each layer, and we use the ReLu and Sigmoid as the activation function at two hidden layers, respectively. The CNN consists of 2 convolution layers with kernel size $3 \times 3$, and 32 kernels at the first layer and 64 kernels at the second layer.  Two fully connected layers with 128 neurons and an output layer with 10 neurons are followed. Cross-entropy function is adopted as the loss function for classification.  We randomly split the entire dataset into training, validation, and testing datasets in a ratio of 5:1:1. 

The results are shown in Tables \ref{mlp_mnist} and \ref{cnn_mnist}. For MLP, we report results for three ANN structures trained by corresponding methods: a) MLP: MLP without adding noises; b) MLP+: MLP with a standard normally distributed noise in each neuron; c) MLPN: MLP with Gaussian noises optimized by our proposed method simultaneously in the process of training synaptic weights by BP. For CNN, we report results for five ANN structures trained by corresponding methods: a) CNN: CNN without adding noises; b) CNN-MLP+: CNN with standard normally distributed noises added only to fully connected neural layers; c) CNN-A+: CNN with standard normally distributed noises added  to both convolution neural layers and fully connected neural layers; d) CNN-MLPN: CNN with Gaussian noises added only to fully connected layers, which are optimized by our proposed method; d) CNN-AN: CNN with Gaussian noises added to both both convolution neural layers and fully connected layers, which are optimized by our proposed method.

\begin{table*}[htb]
\begin{minipage}{\columnwidth}
\small
\centering
\begin{tabular}{|c|l|l|l|l|l|l|l|l|l|l|l|}
\hline
\multirow{2}{*}{Models} &
  \multicolumn{1}{c|}{\multirow{2}{*}{Act.}} &
  \multicolumn{1}{c|}{\multirow{2}{*}{Ori}} &
  \multicolumn{3}{c|}{White box    evaluation} &
  \multicolumn{6}{c|}{Black box  evaluation} \\ \cline{4-12} 
 &
  \multicolumn{1}{c|}{} &
  \multicolumn{1}{c|}{} &
  FGSM &
  L-BFGS &
  PGD &
  Gaussian &
  Impluse &
  Glass Blur &
  Contrast &
  FGSM &
  L-BFGS \\ \hline
\multirow{2}{*}{MLP}  & ReLU    & 0.948 & 0.149 & 0.255 & 0.105 & 0.935 & 0.934 & 0.879 & 0.598 & 0.314 & 0.690 \\ \cline{2-12} 
                      & Sigmoid & 0.936 & 0.280 & 0.324 & 0.207 & 0.883 & 0.783 & 0.885 & 0.676 & 0.410 & 0.749 \\ \hline
\multirow{2}{*}{MLP+} & ReLU    & 0.884 & 0.283 & 0.413 & 0.238 & 0.875 & 0.851 & 0.531 & 0.531 & 0.340 & 0.742 \\ \cline{2-12} 
                      & Sigmoid & 0.875 & 0.314 & 0.433 & 0.253 & 0.869 & 0.834 & 0.817 & 0.605 & 0.432 & 0.736 \\ \hline
\multirow{2}{*}{MLPN} & ReLU    & 0.921 & 0.295 & 0.420 & 0.267 & 0.895 & 0.909 & 0.835 & 0.672 & 0.430 & 0.745 \\ \cline{2-12} 
 &
  Sigmoid &
  \textbf{0.957} &
  \textbf{0.336} &
  \textbf{0.568} &
  \textbf{0.275} &
  \textbf{0.946} &
  \textbf{0.944} &
  \textbf{0.920} &
  \textbf{0.710} &
  \textbf{0.465} &
  \textbf{0.788} \\ \hline
\end{tabular}
\caption{Evaluations results of MLP models in MNIST}
\label{mlp_mnist}
\end{minipage}

\begin{minipage}{\columnwidth}
\small
\centering
\begin{tabular}{|l|l|l|l|l|l|l|l|l|l|l|}
\hline
\multicolumn{1}{|c|}{\multirow{2}{*}{Models}} &
  \multicolumn{1}{c|}{\multirow{2}{*}{Ori}} &
  \multicolumn{3}{c|}{White box  evaluation} &
  \multicolumn{6}{c|}{Black box evaluation} \\ \cline{3-11} 
\multicolumn{1}{|c|}{} & \multicolumn{1}{c|}{} & FGSM  & L-BFGS         & PGD   & Gaussian       & Impluse & Glass Blur     & Contrast & FGSM  & L-BFGS         \\ \hline
CNN                    & 0.986                 & 0.744 & 0.616          & 0.655 & 0.983 & 0.971   & 0.752          & 0.845    & 0.917 & 0.779          \\ \hline
CNN-MLP+               & 0.980                  & 0.788 & 0.613          & 0.684 & 0.977          & 0.955   & 0.564          & 0.794    & 0.924 & 0.767          \\ \hline
CNN-A+                 & 0.974                 & 0.757 & 0.586          & 0.704 & 0.951          & 0.947   & 0.835          & 0.575    & 0.920 & 0.775          \\ \hline
CNN-MLPN &
  \textbf{0.990} &
  \textbf{0.870} &
  0.685 &
  \textbf{0.752} &
  \textbf{0.995} &
  \textbf{0.984} &
  0.788 &
  \textbf{0.853} &
  \textbf{0.957} &
  0.818 \\ \hline
CNN-AN                 & 0.982                 & 0.783 & \textbf{0.766} & 0.714 & 0.976          & 0.973   & \textbf{0.867} & 0.834    & 0.928 & \textbf{0.826} \\ \hline
\end{tabular}
\caption{Evaluation results of CNN models  in MNIST}
\label{cnn_mnist}
\end{minipage}

\end{table*}

\begin{figure*}[h] 
\small
\centering
\begin{minipage}[t]{0.35\linewidth} 
\includegraphics[scale=0.25]{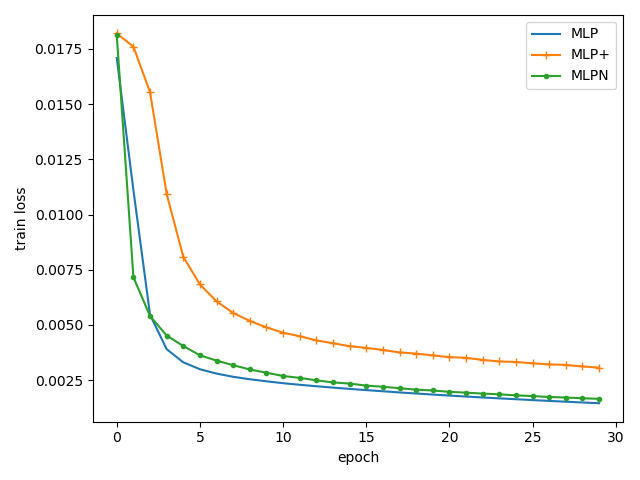} 
\label{trainloss} 
\end{minipage}%
\begin{minipage}[t]{0.35\linewidth} 
\includegraphics[scale=0.25]{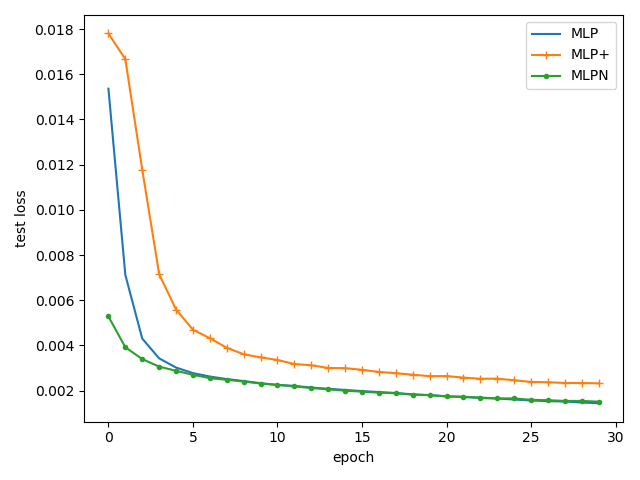} 
\label{testloss} 
\end{minipage}%
\begin{minipage}[t]{0.35\linewidth} 
\includegraphics[scale=0.25]{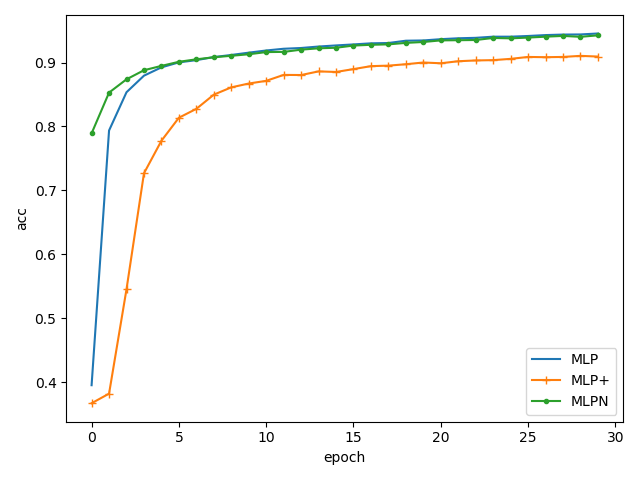} 
\label{acc}
\end{minipage}
\caption{The figures from left to right respectively show the training loss, validation loss and accuracy in testing dataset.}
\label{accelerationOnMLP}
\end{figure*}

\begin{figure*}[ht] 
\small
\begin{minipage}[t]{0.16\linewidth} 
\centering 
\includegraphics[width=1.3in]{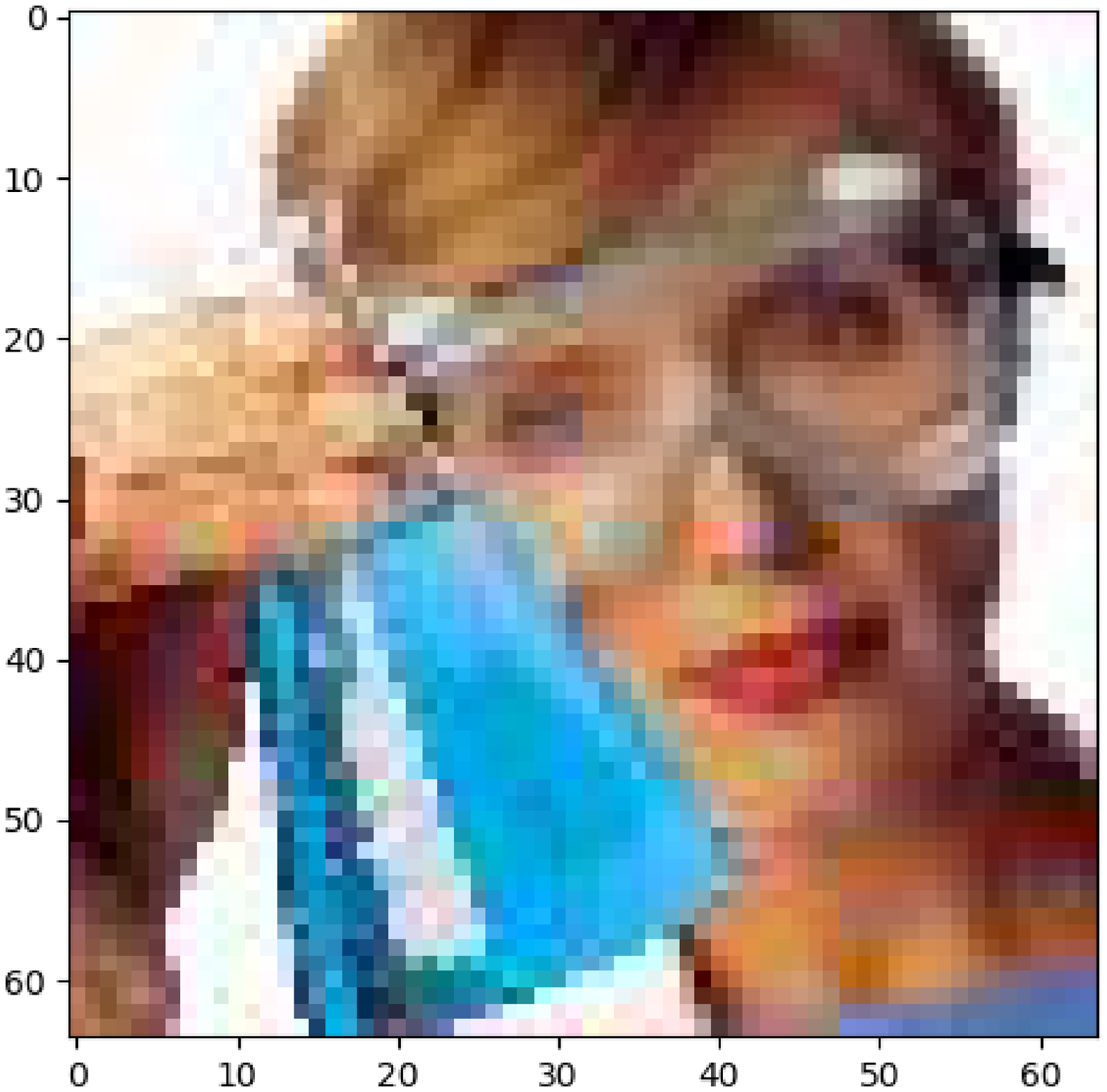} 
\label{person} 
\end{minipage}%
\begin{minipage}[t]{0.16\linewidth} 
\centering 
\includegraphics[width=1.3in]{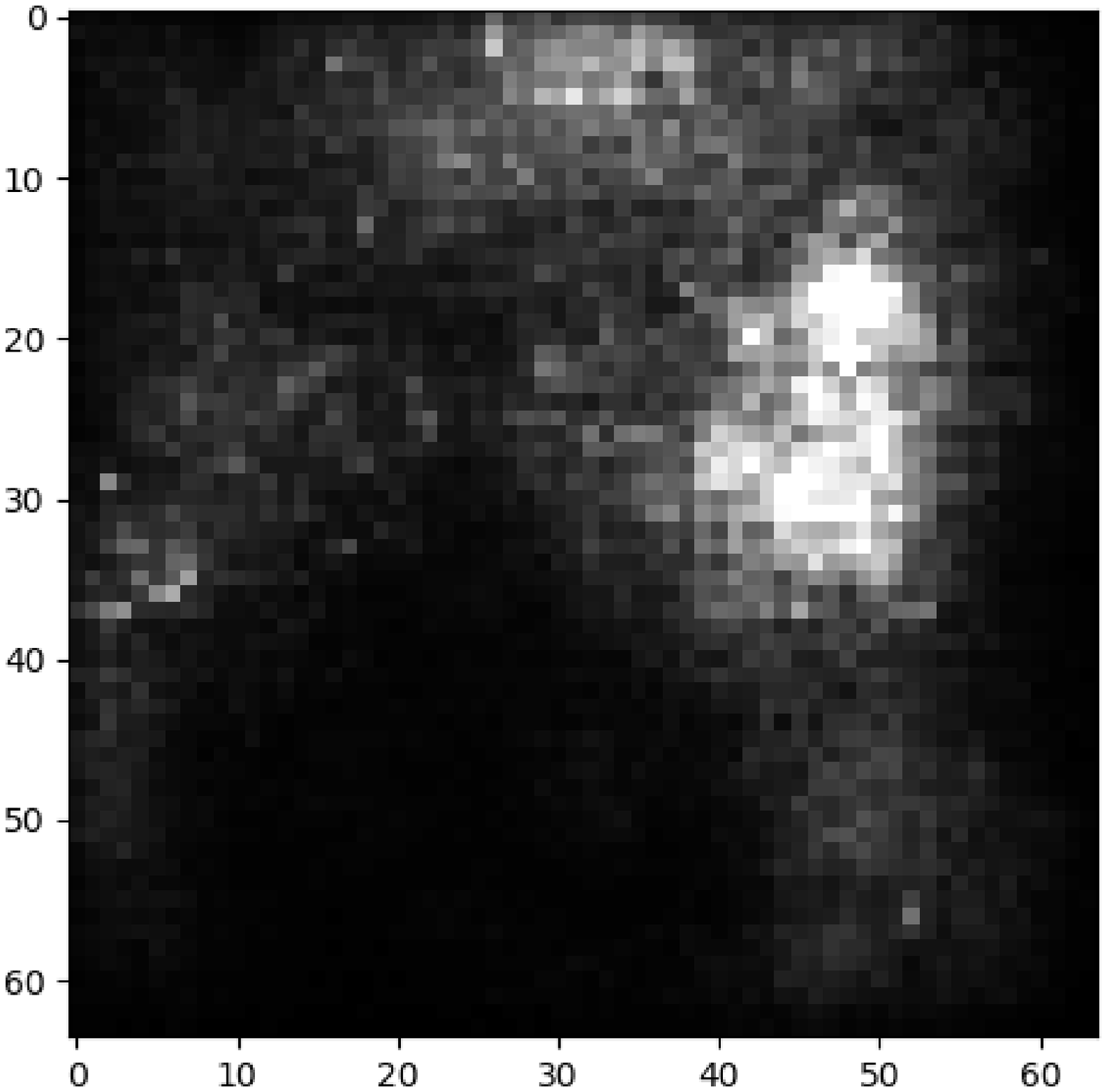} 
\label{person1} 
\end{minipage}%
\begin{minipage}[t]{0.16\linewidth} 
\centering 
\includegraphics[width=1.3in]{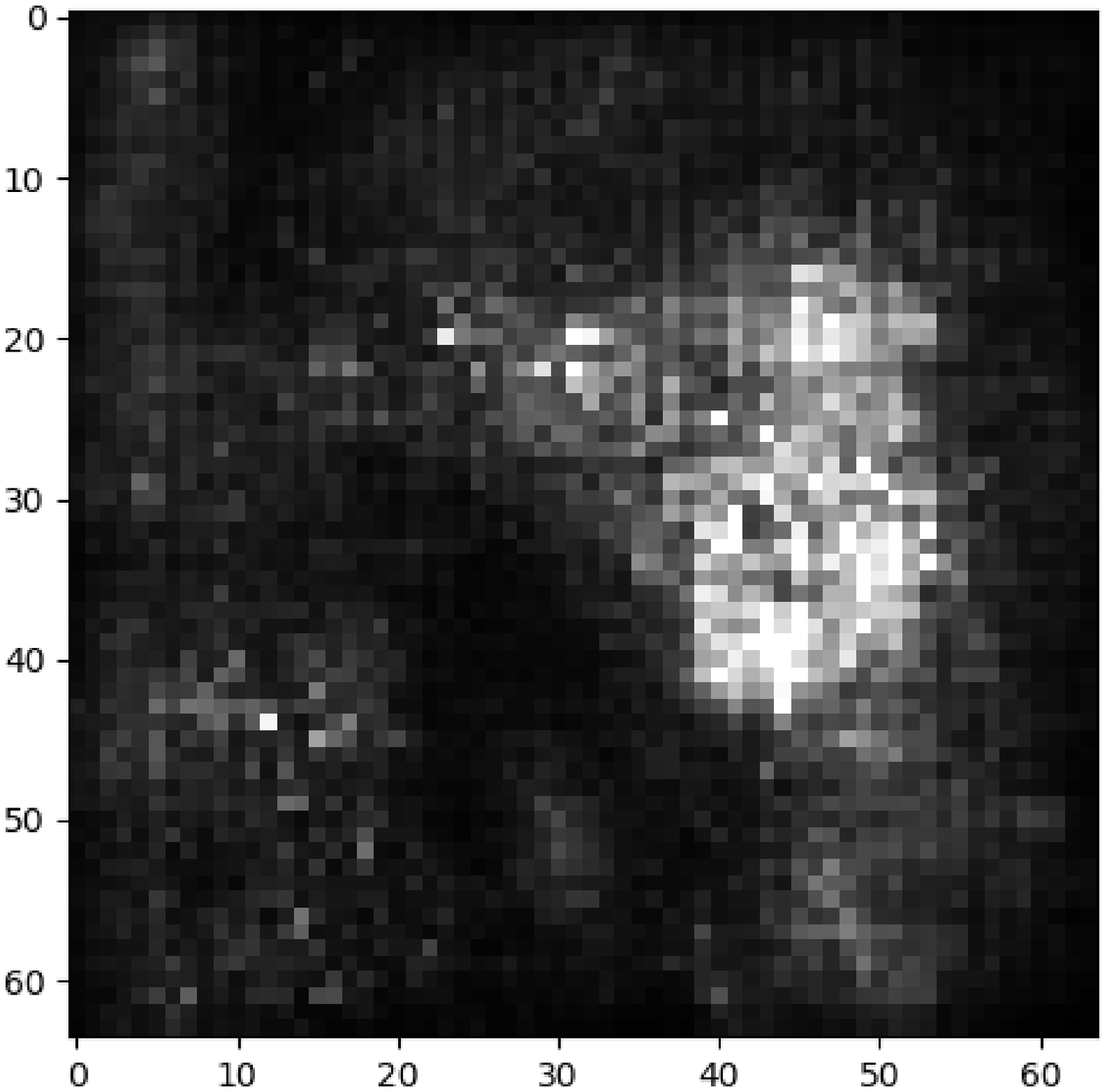} 
\label{person2} 
\end{minipage}%
\begin{minipage}[t]{0.16\linewidth} 
\centering 
\includegraphics[width=1.3in]{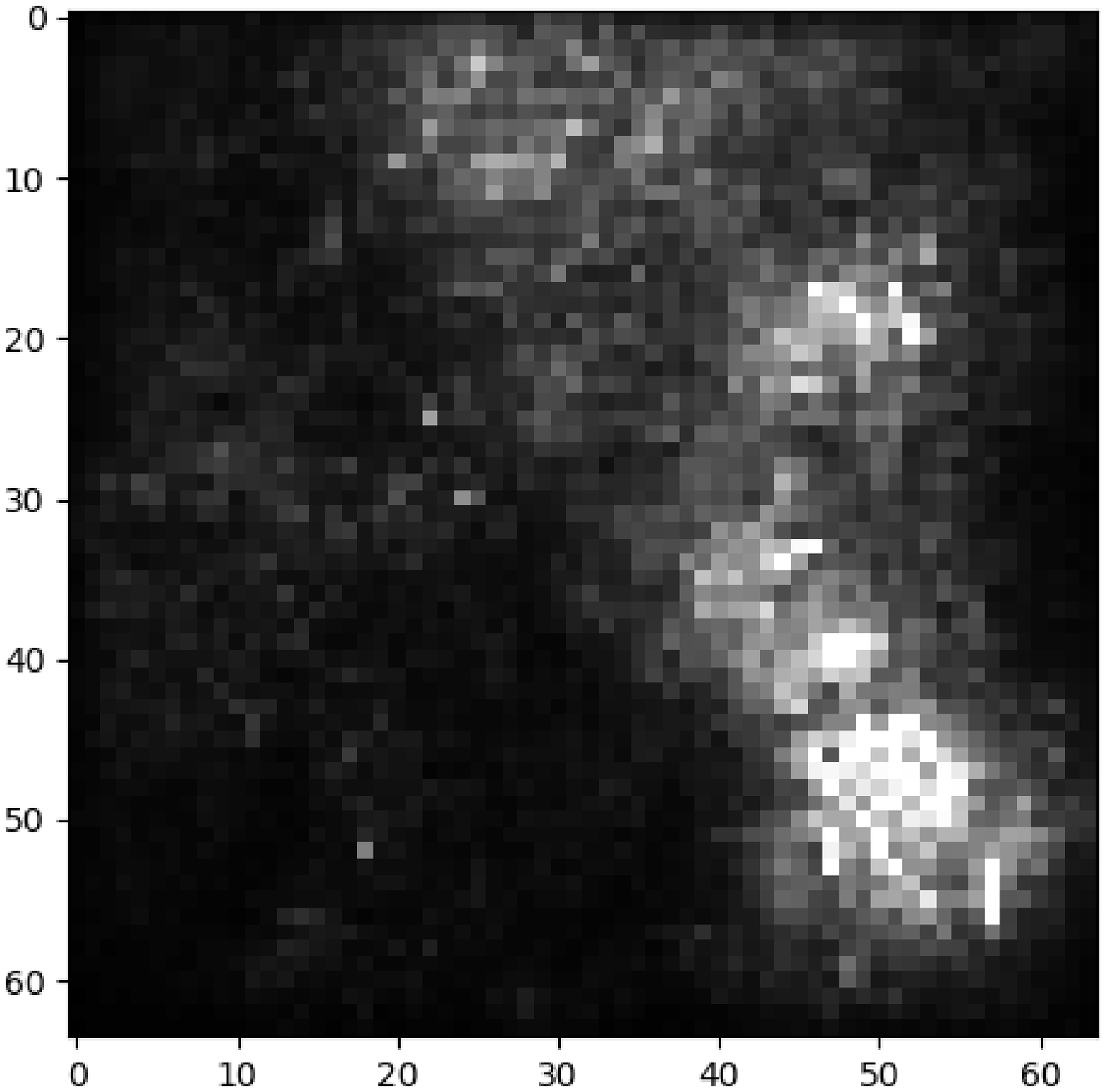} 
\label{person4} 
\end{minipage}%
\begin{minipage}[t]{0.16\linewidth} 
\centering 
\includegraphics[width=1.3in]{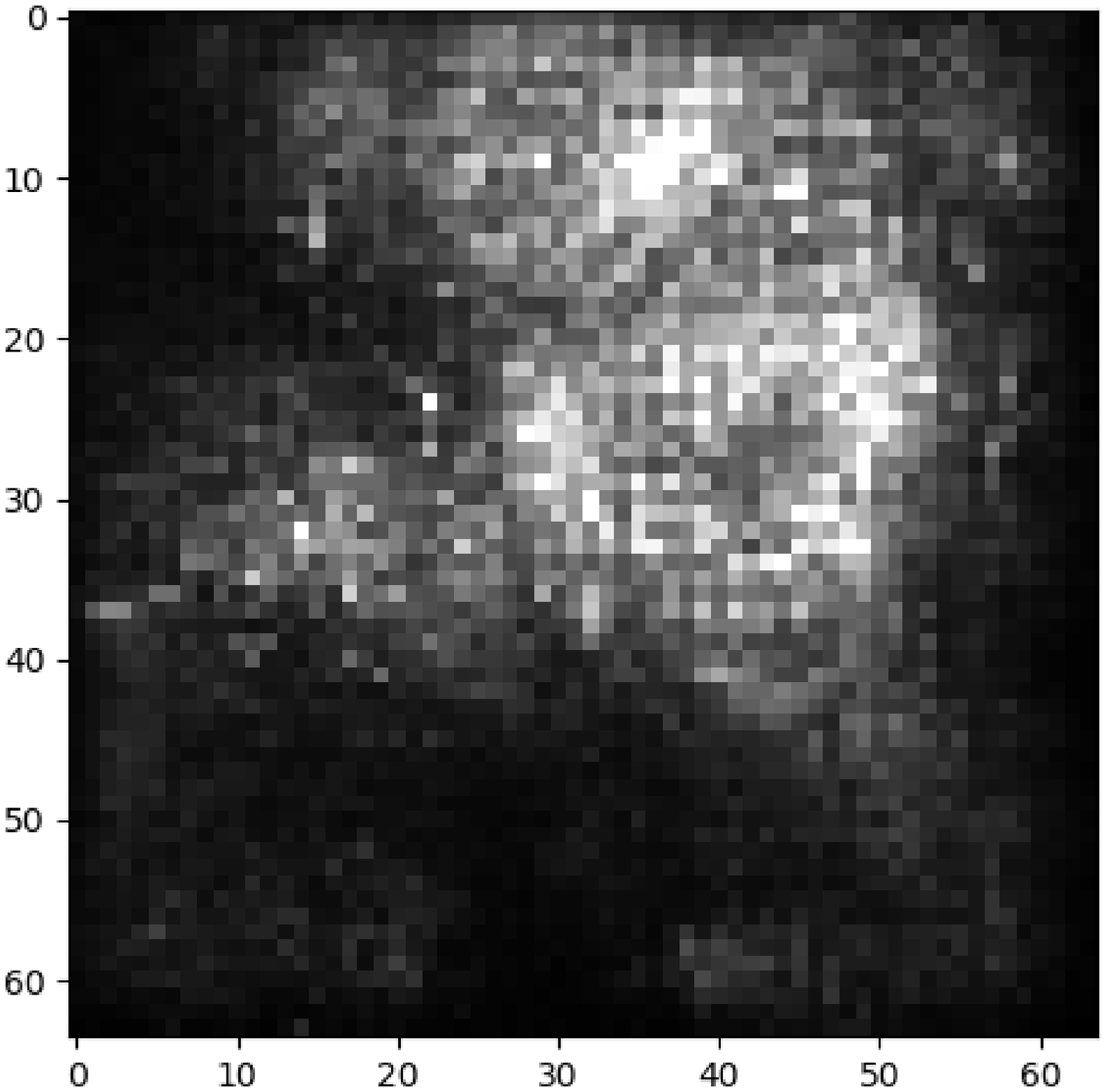} 
\label{person3} 
\end{minipage}%
\begin{minipage}[t]{0.16\linewidth} 
\centering 
\includegraphics[width=1.3in]{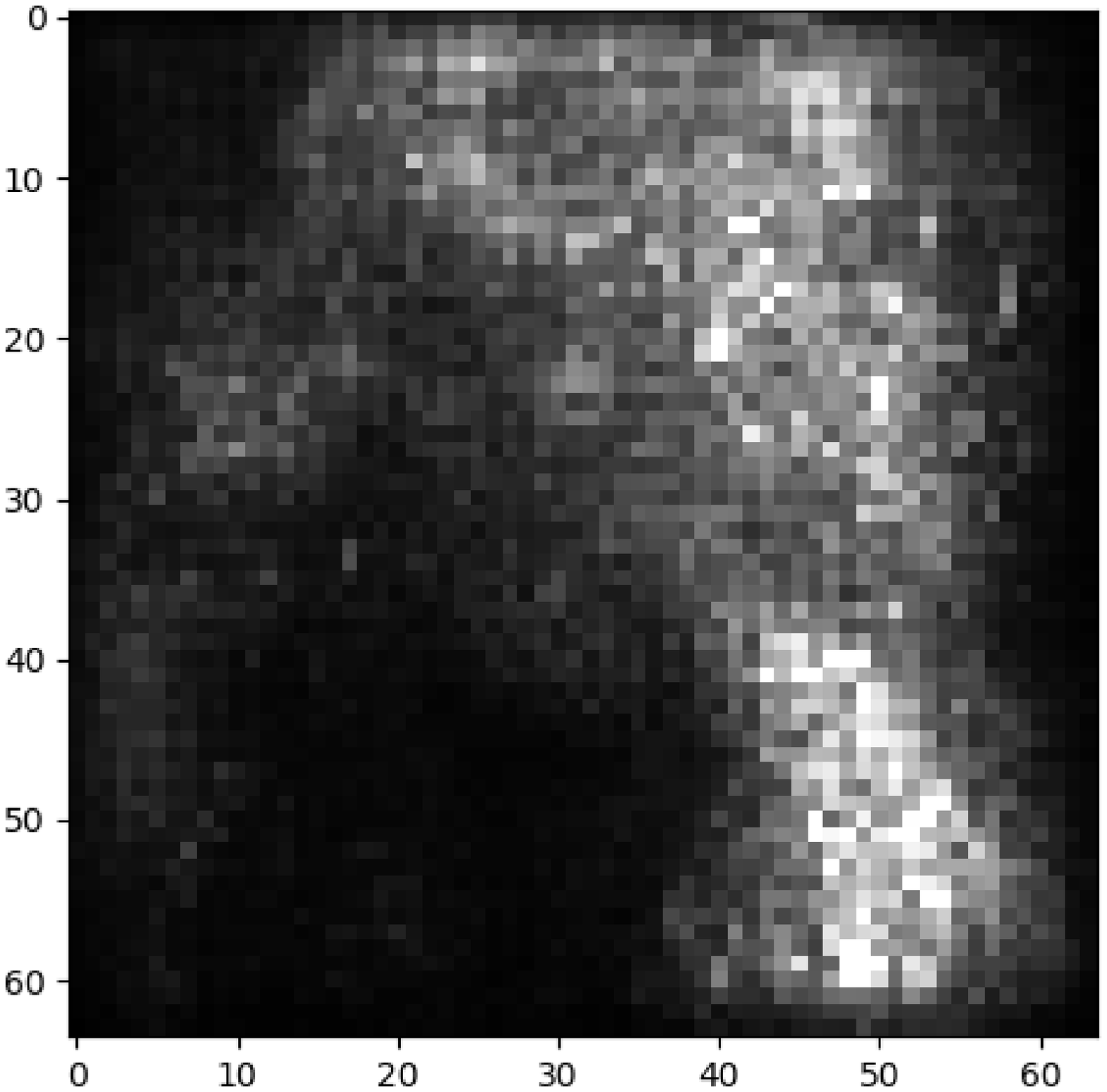} 
\label{person5} 
\end{minipage}%

\begin{minipage}[t]{0.16\linewidth} 
\centering 
\includegraphics[width=1.3in]{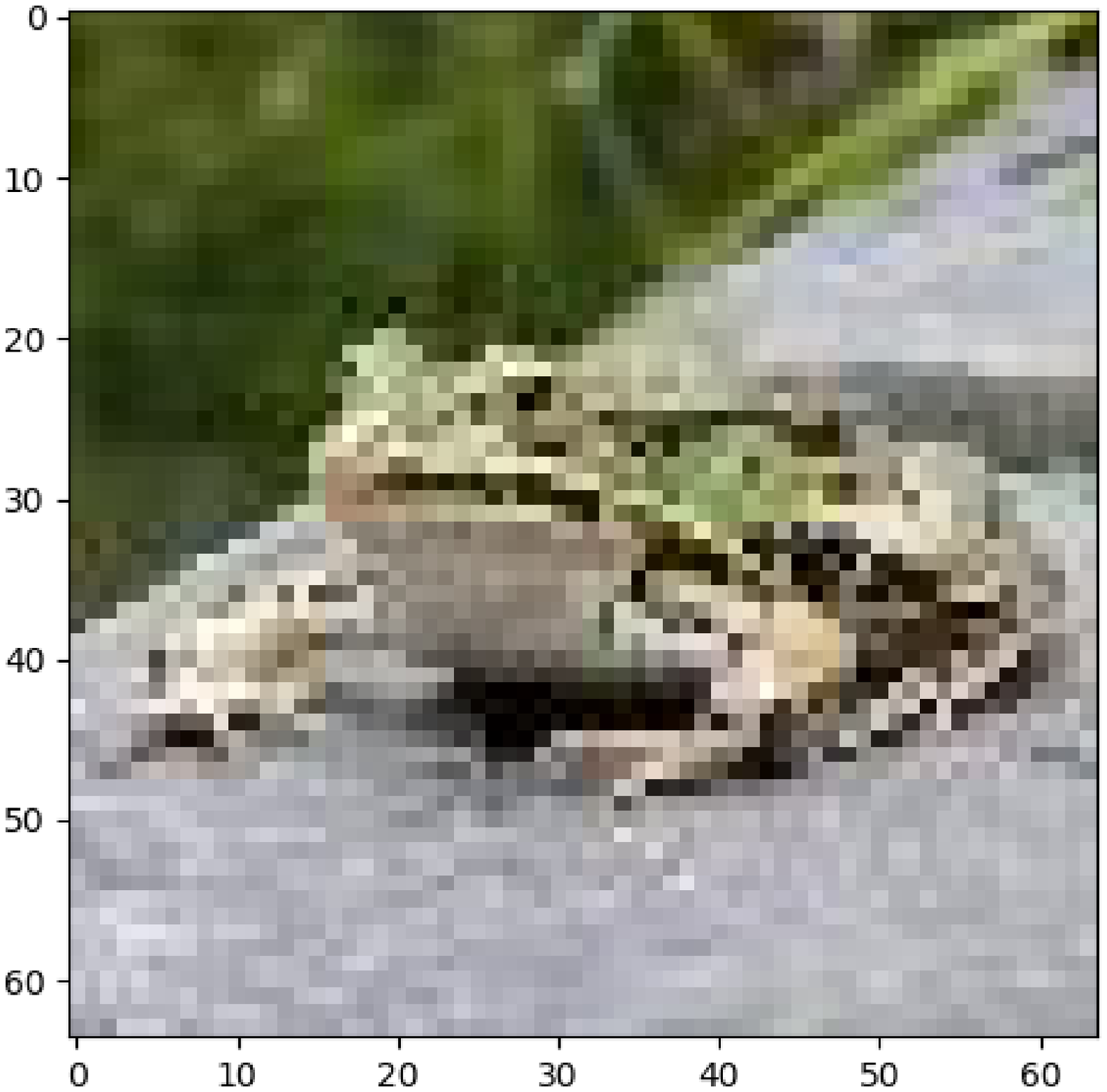} 
\label{frog} 
\end{minipage}%
\begin{minipage}[t]{0.16\linewidth} 
\centering 
\includegraphics[width=1.3in]{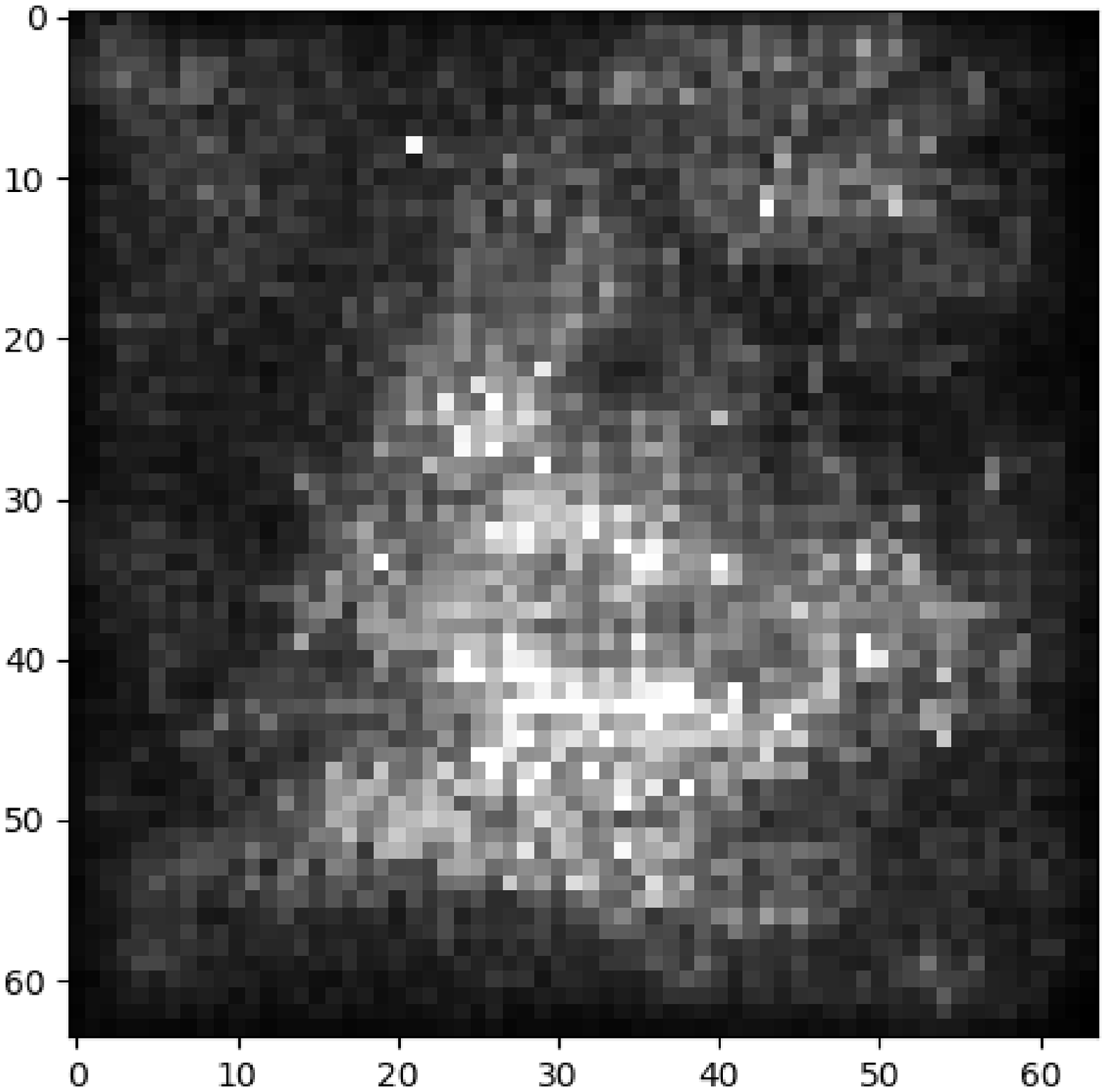} 
\label{frog1} 
\end{minipage}%
\begin{minipage}[t]{0.16\linewidth} 
\centering 
\includegraphics[width=1.3in]{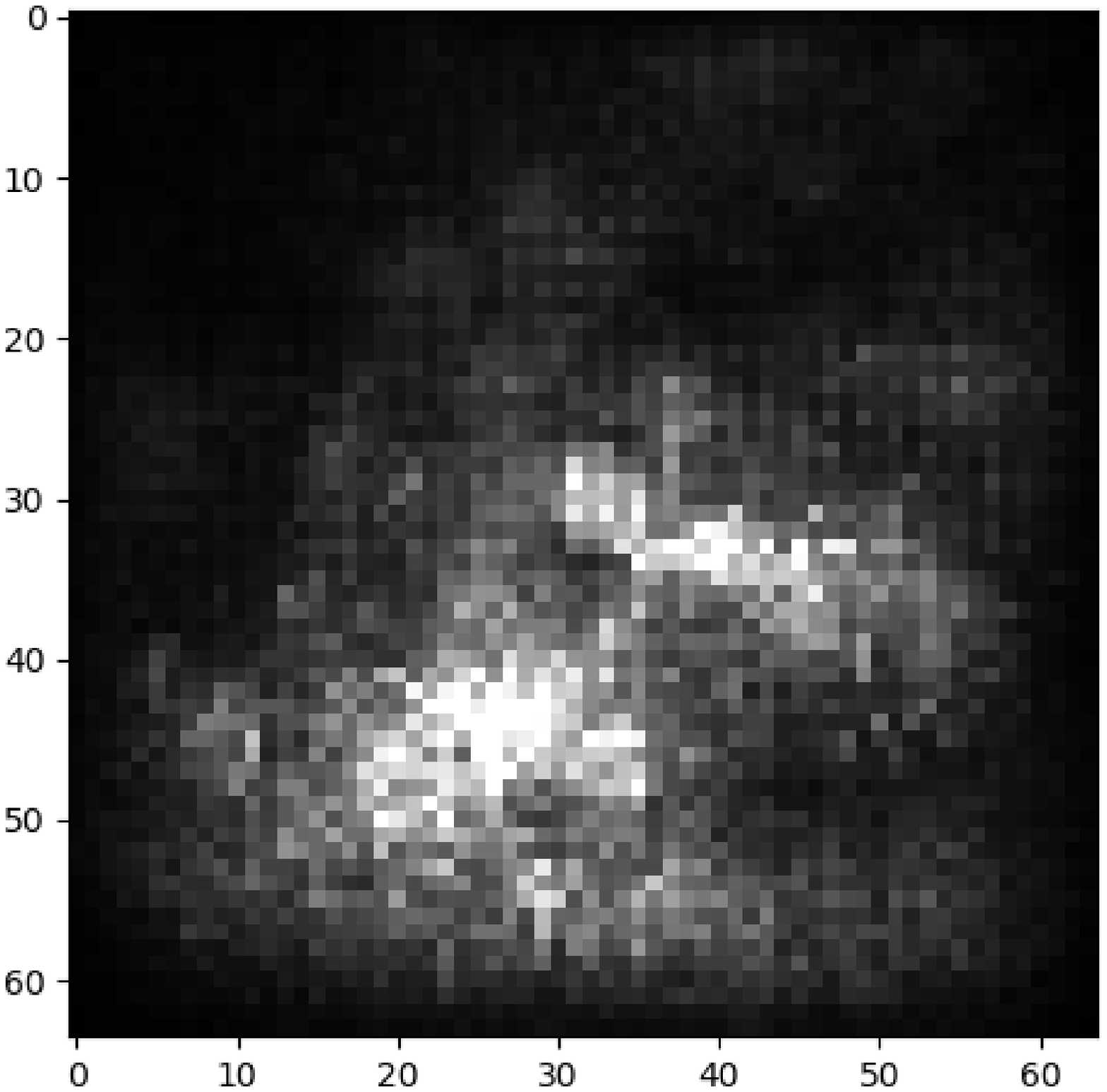} 
\label{frog2} 
\end{minipage}%
\begin{minipage}[t]{0.16\linewidth} 
\centering 
\includegraphics[width=1.3in]{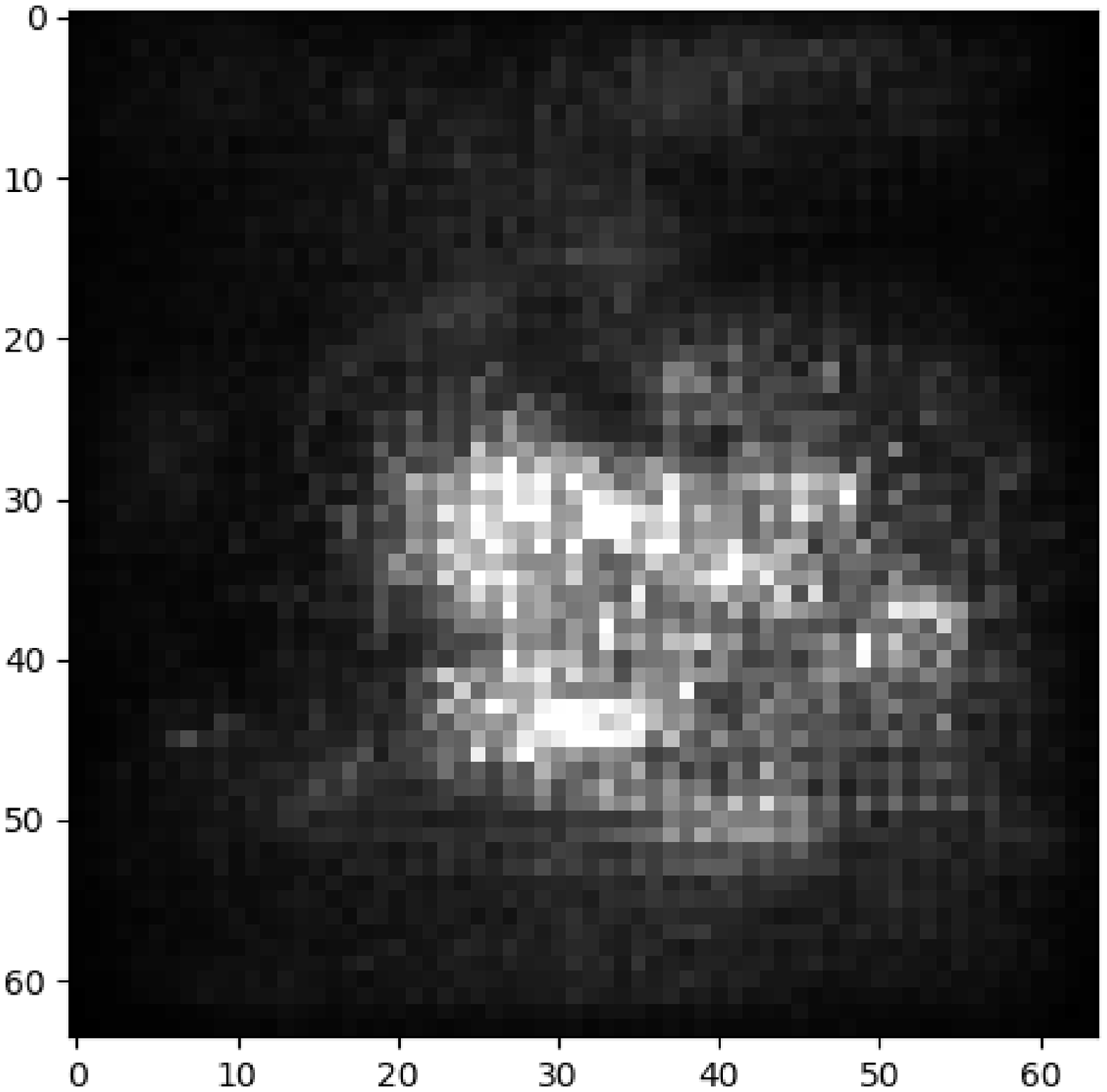} 
\label{frog4} 
\end{minipage}%
\begin{minipage}[t]{0.16\linewidth} 
\centering 
\includegraphics[width=1.3in]{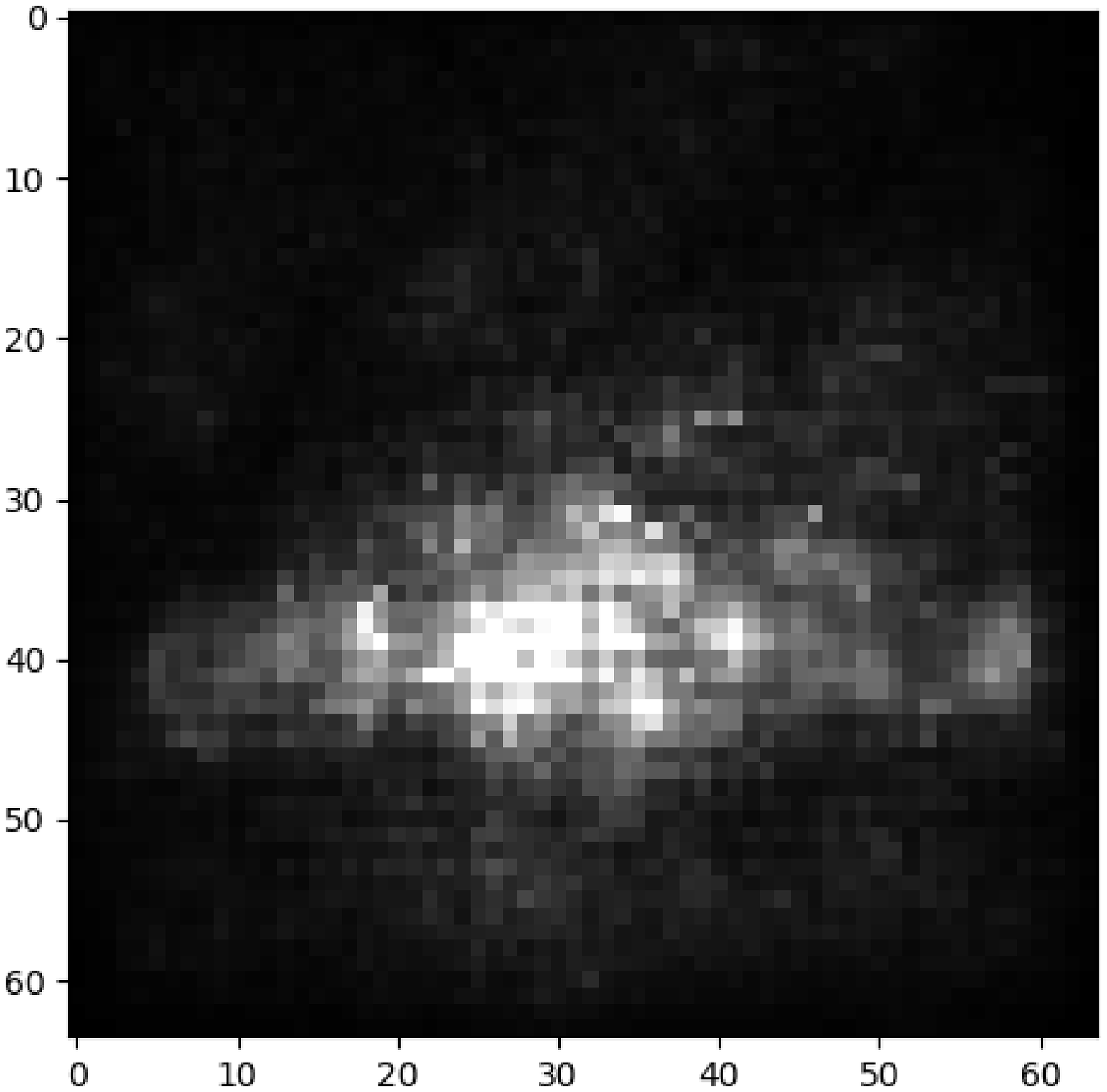} 
\label{frog3} 
\end{minipage}%
\begin{minipage}[t]{0.16\linewidth} 
\centering 
\includegraphics[width=1.3in]{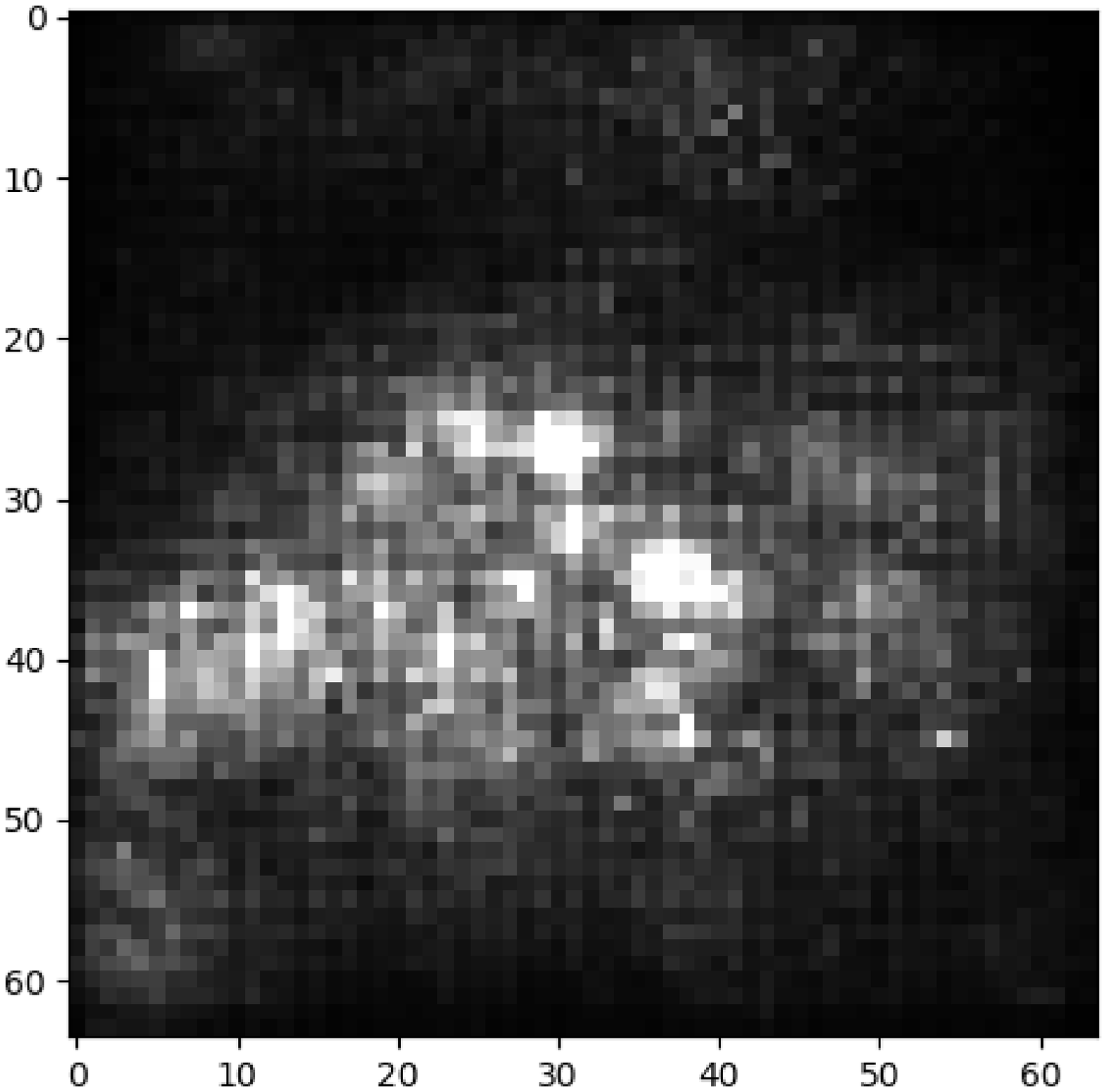} 
\label{frog5} 
\end{minipage}%

\begin{minipage}[t]{0.16\linewidth} 
\centering 
\includegraphics[width=1.3in]{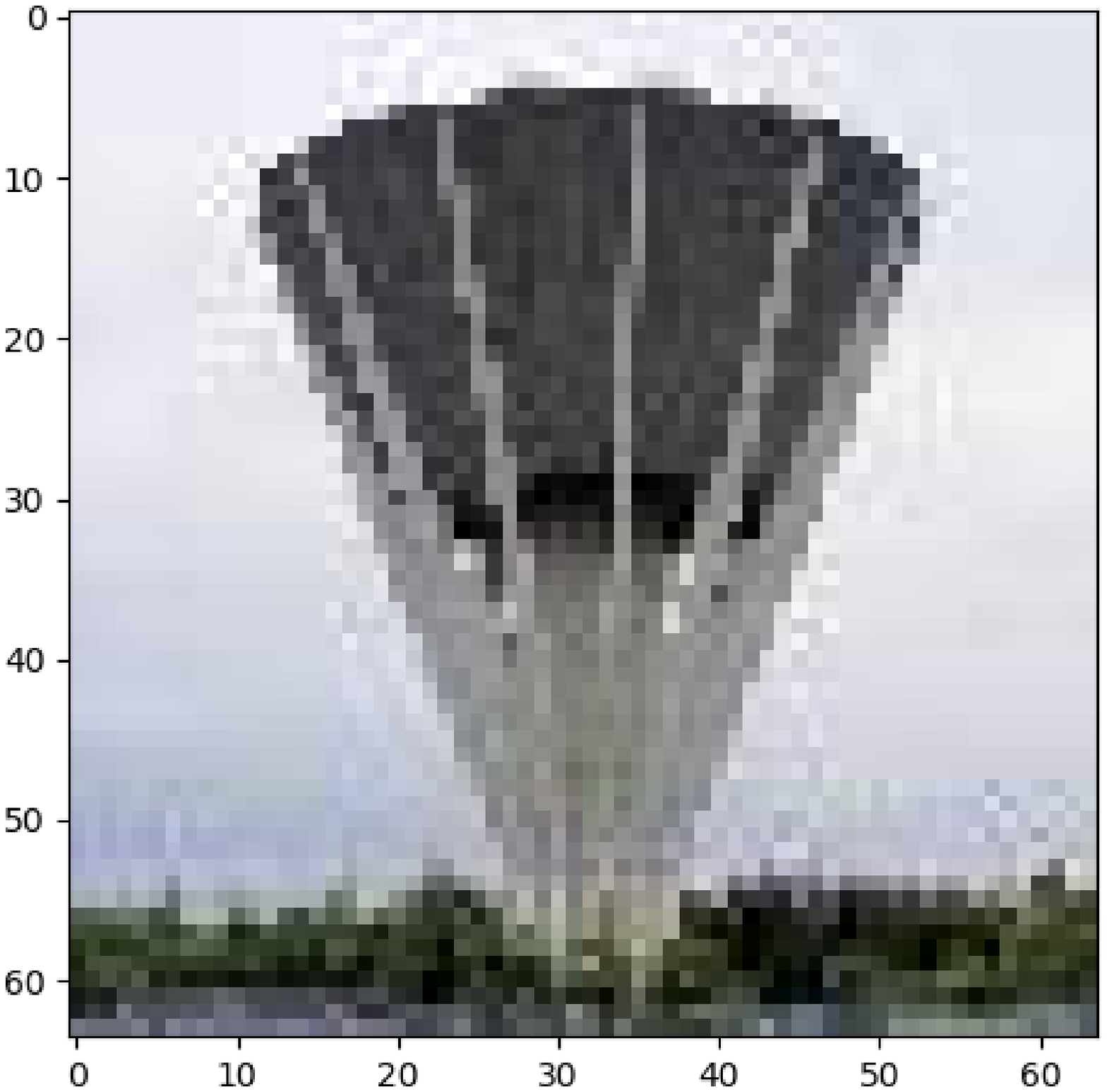} 
\label{zhuzi} 
\end{minipage}%
\begin{minipage}[t]{0.16\linewidth} 
\centering 
\includegraphics[width=1.3in]{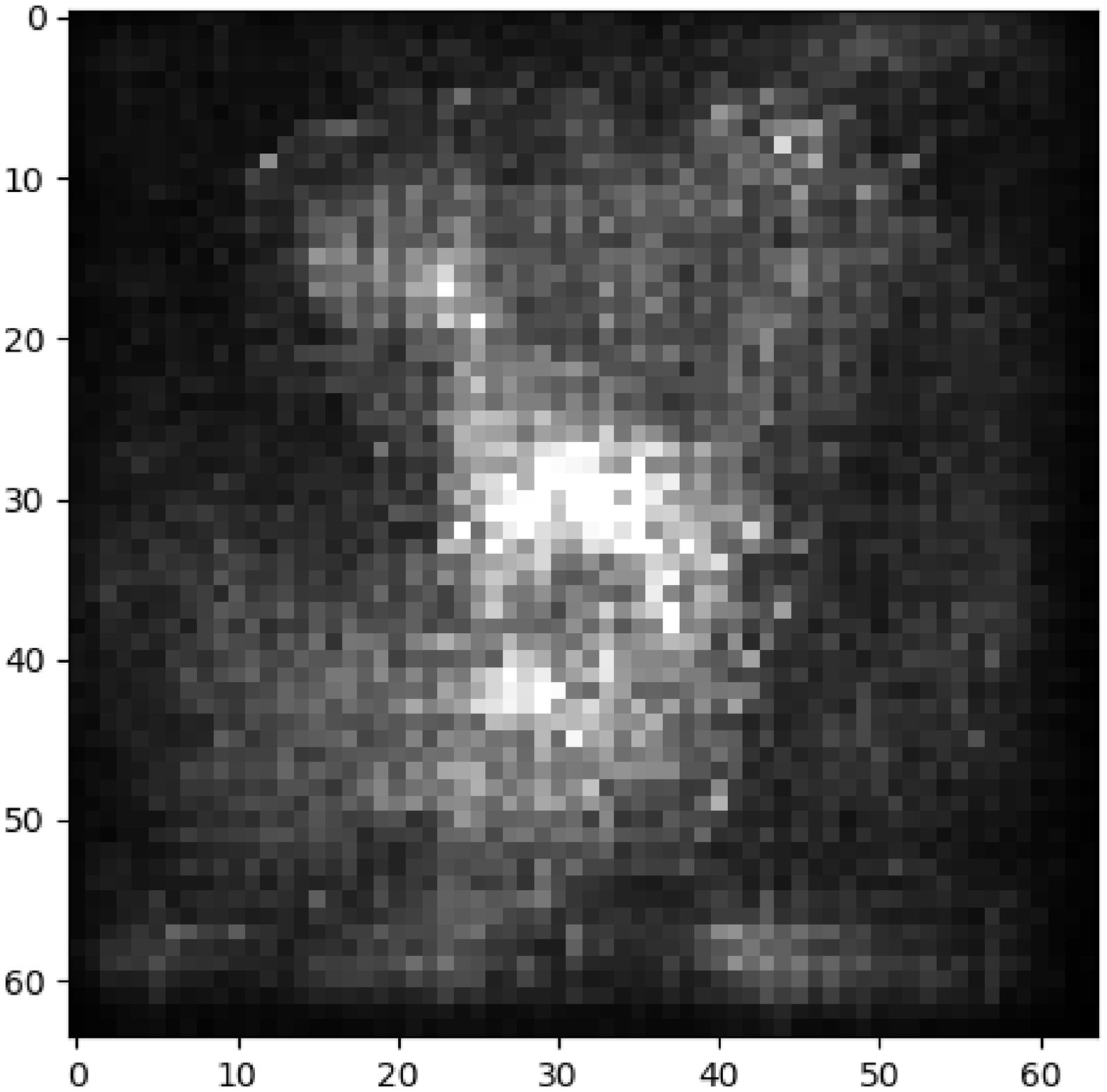} 
\label{zhuzi1} 
\end{minipage}%
\begin{minipage}[t]{0.16\linewidth} 
\centering 
\includegraphics[width=1.3in]{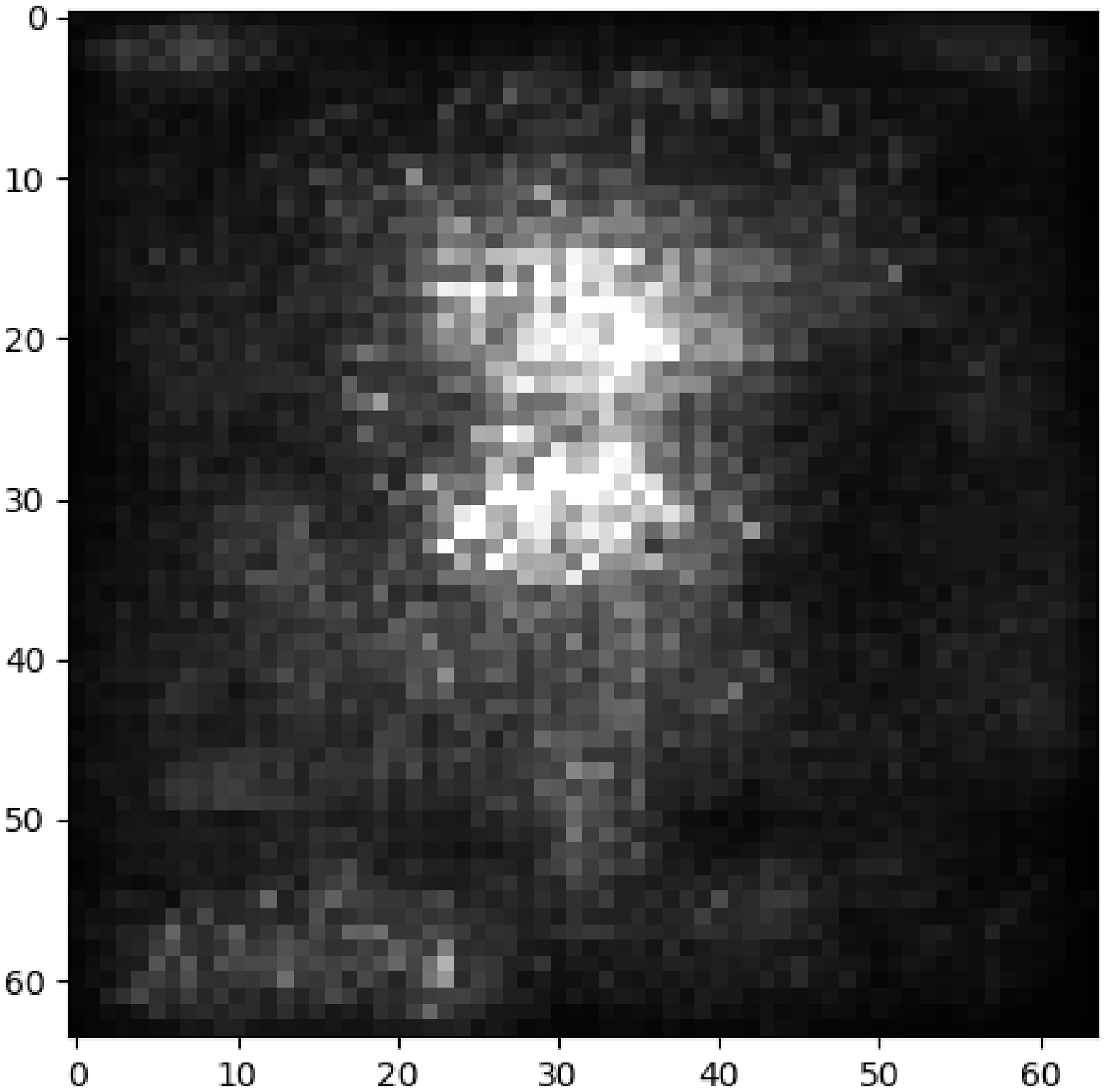} 
\label{zhuzi2} 
\end{minipage}%
\begin{minipage}[t]{0.16\linewidth} 
\centering 
\includegraphics[width=1.3in]{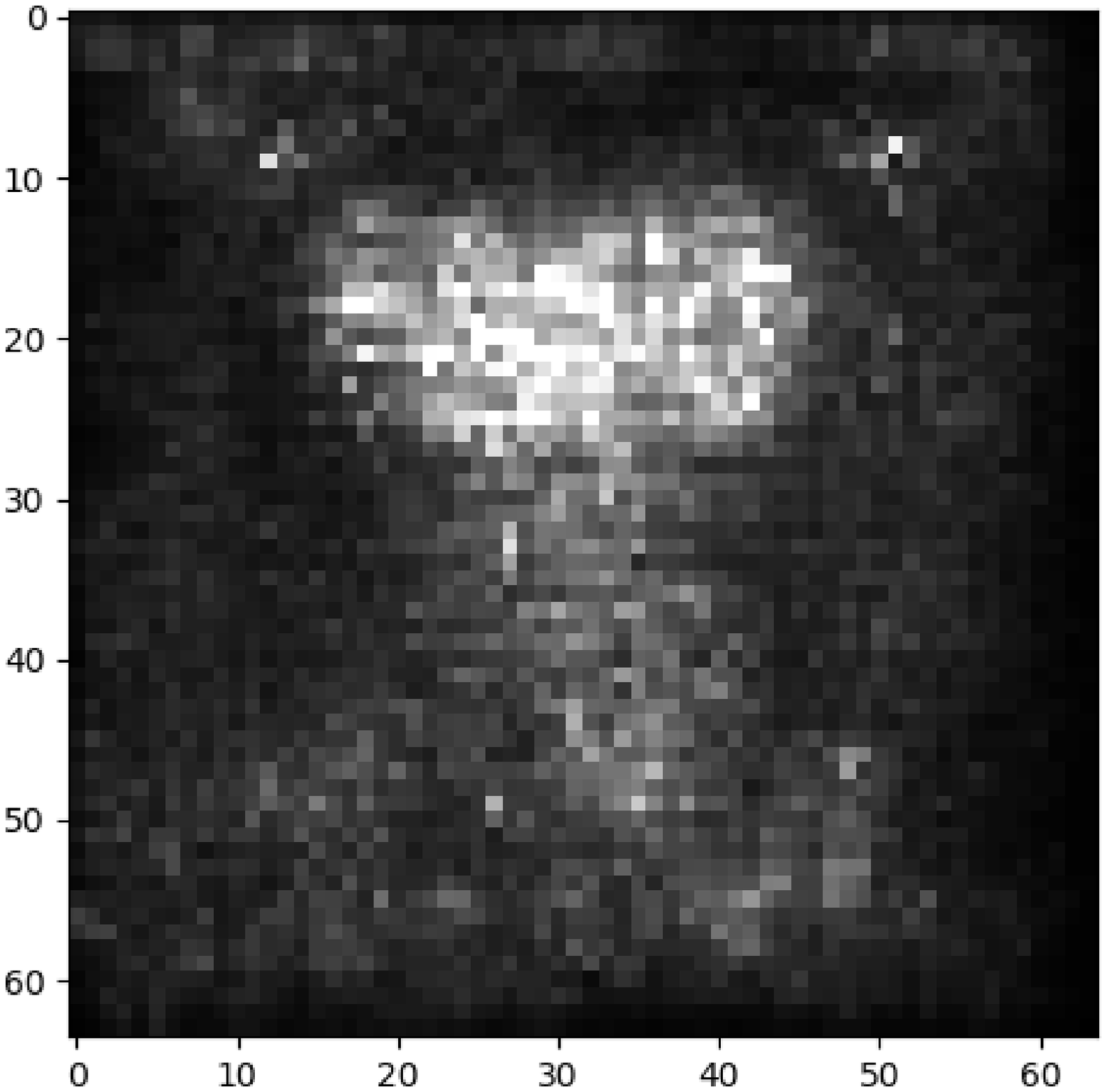} 
\label{zhuzi4}
\end{minipage}%
\begin{minipage}[t]{0.16\linewidth} 
\centering 
\includegraphics[width=1.3in]{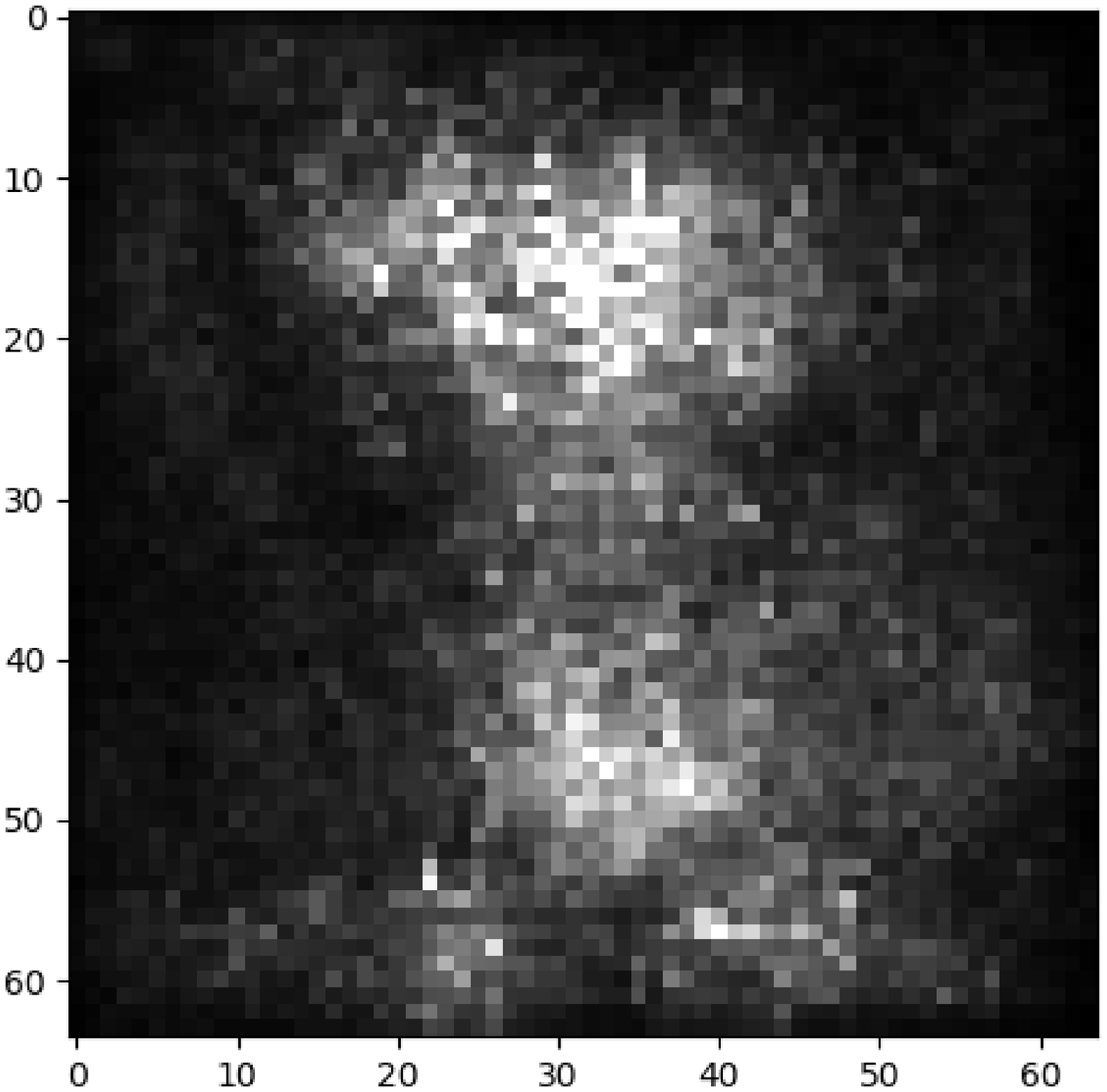} 
\label{zhuzi3} 
\end{minipage}%
\begin{minipage}[t]{0.16\linewidth} 
\centering 
\includegraphics[width=1.3in]{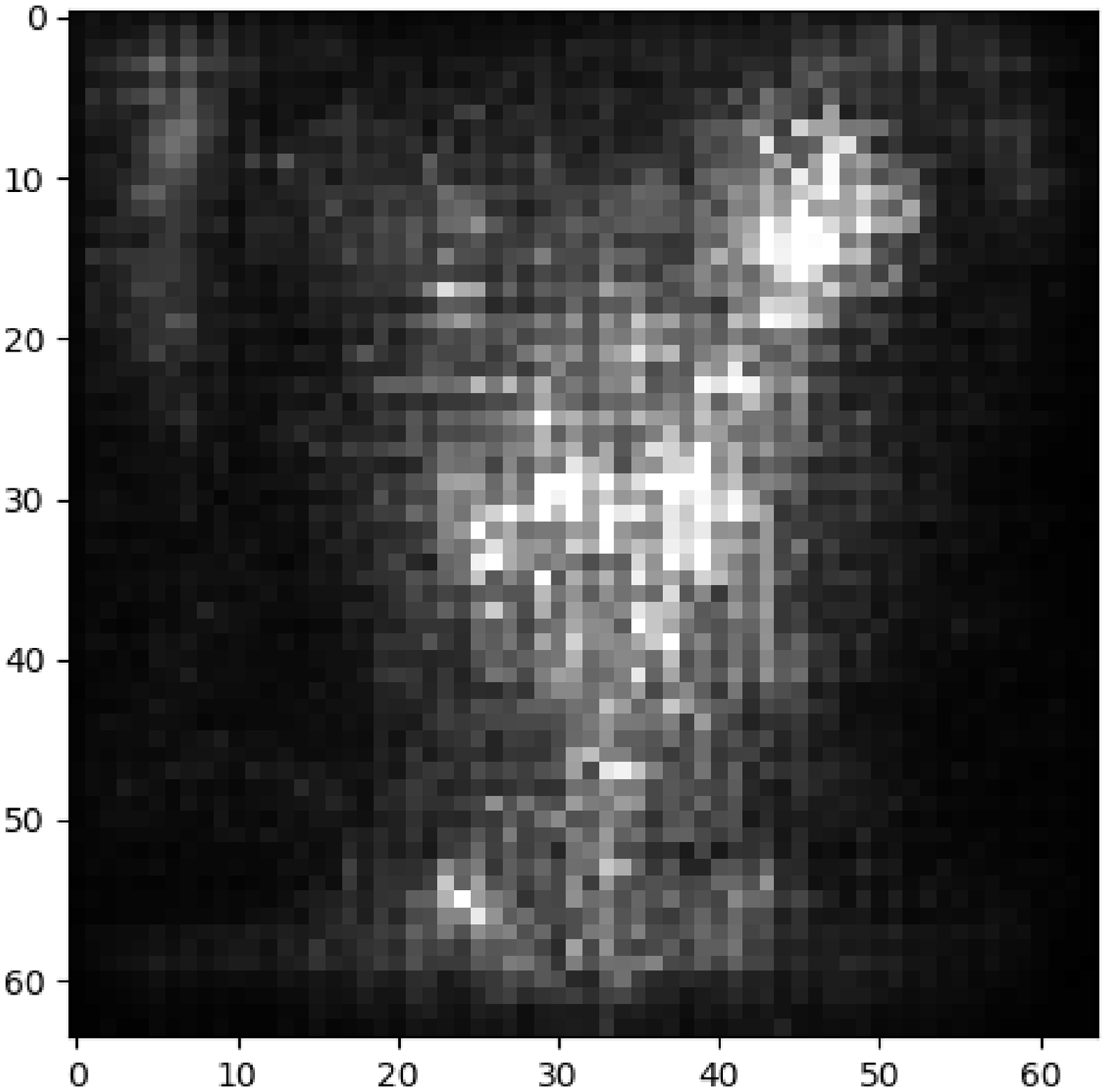} 
\label{zhuzi5} 
\end{minipage}%

\begin{minipage}[t]{0.16\linewidth} 
\centering 
\includegraphics[width=1.3in]{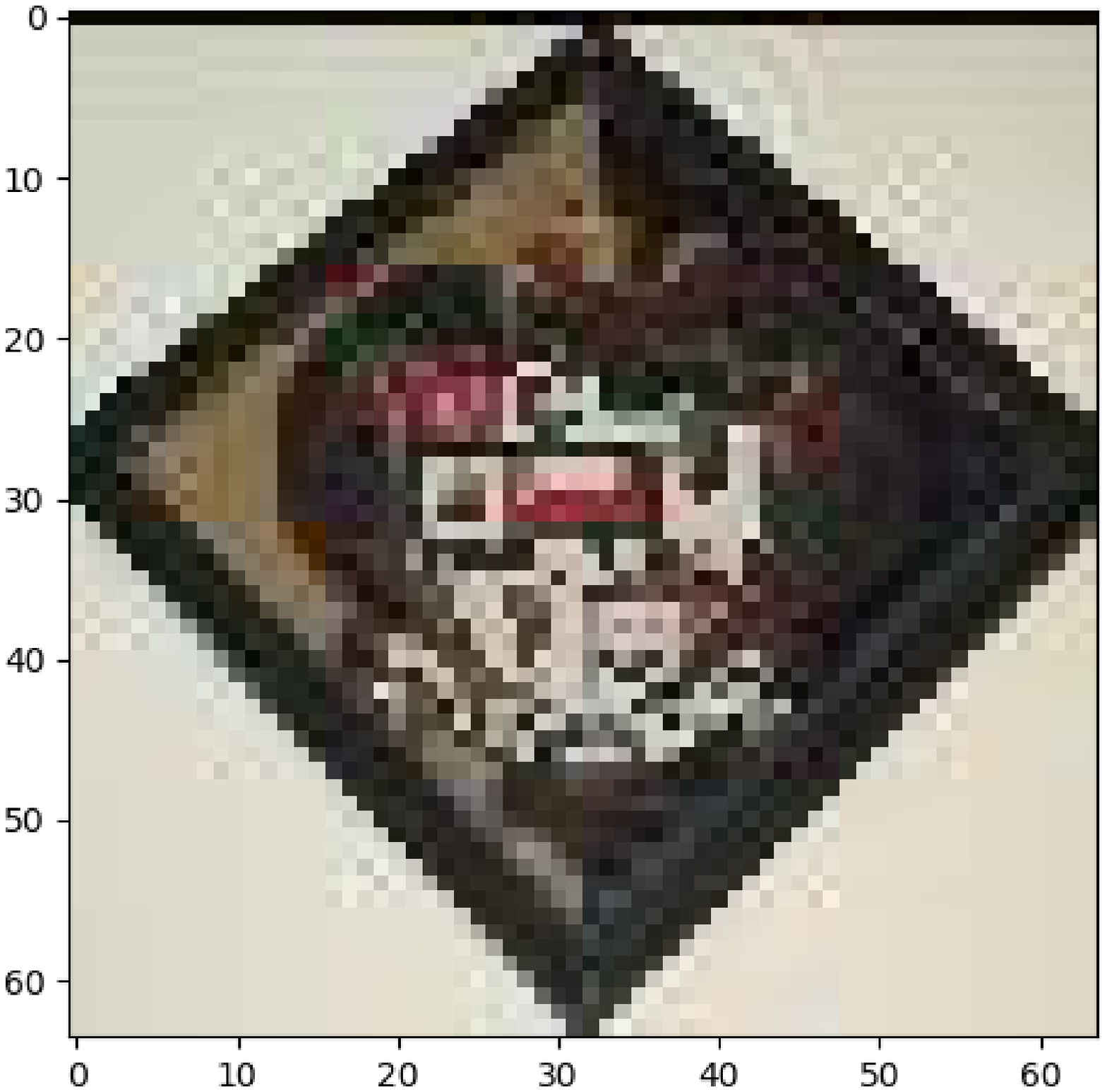} 
\label{fu} 
\end{minipage}%
\begin{minipage}[t]{0.16\linewidth} 
\centering 
\includegraphics[width=1.3in]{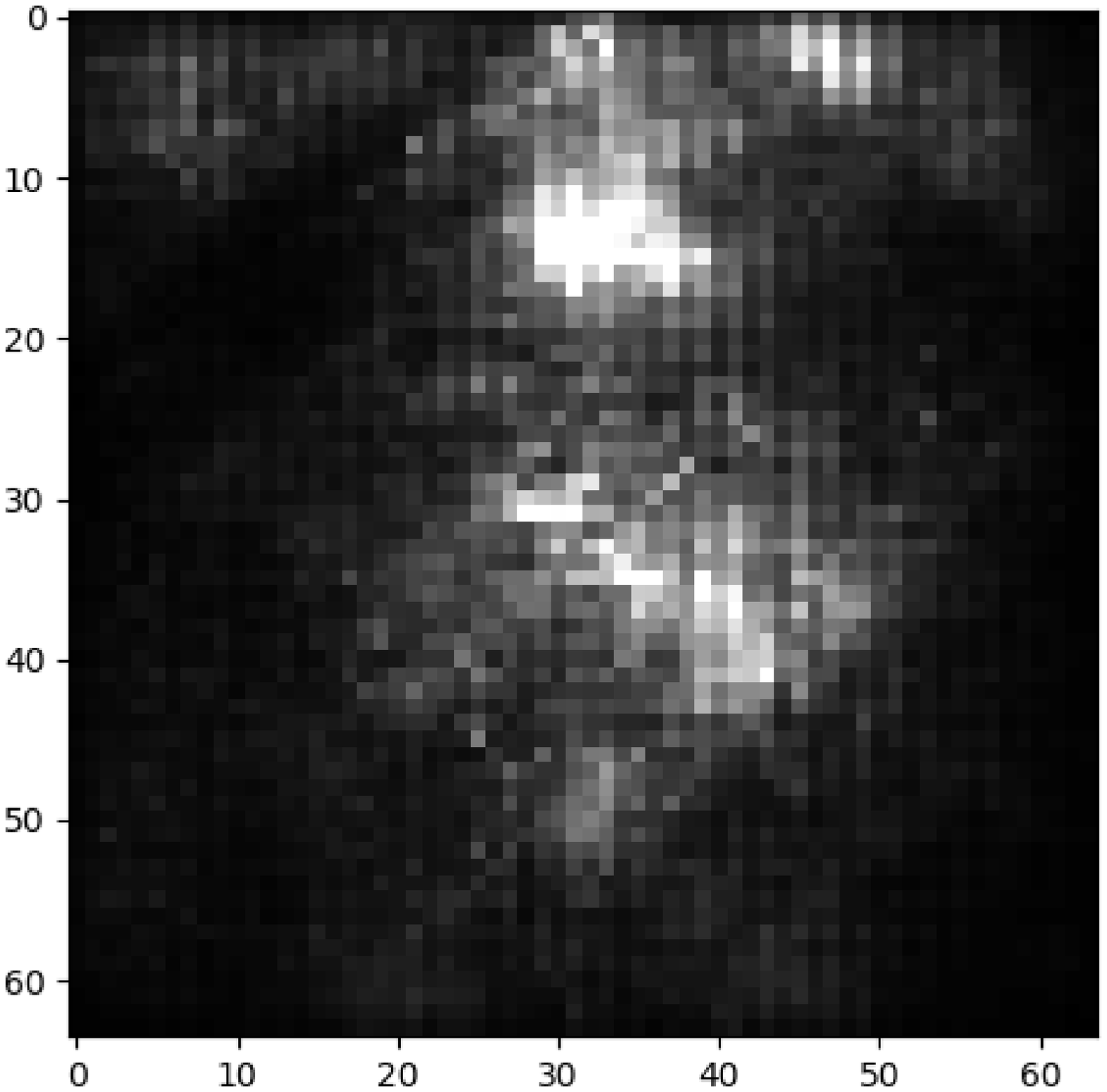} 
\label{fu1} 
\end{minipage}%
\begin{minipage}[t]{0.16\linewidth} 
\centering 
\includegraphics[width=1.3in]{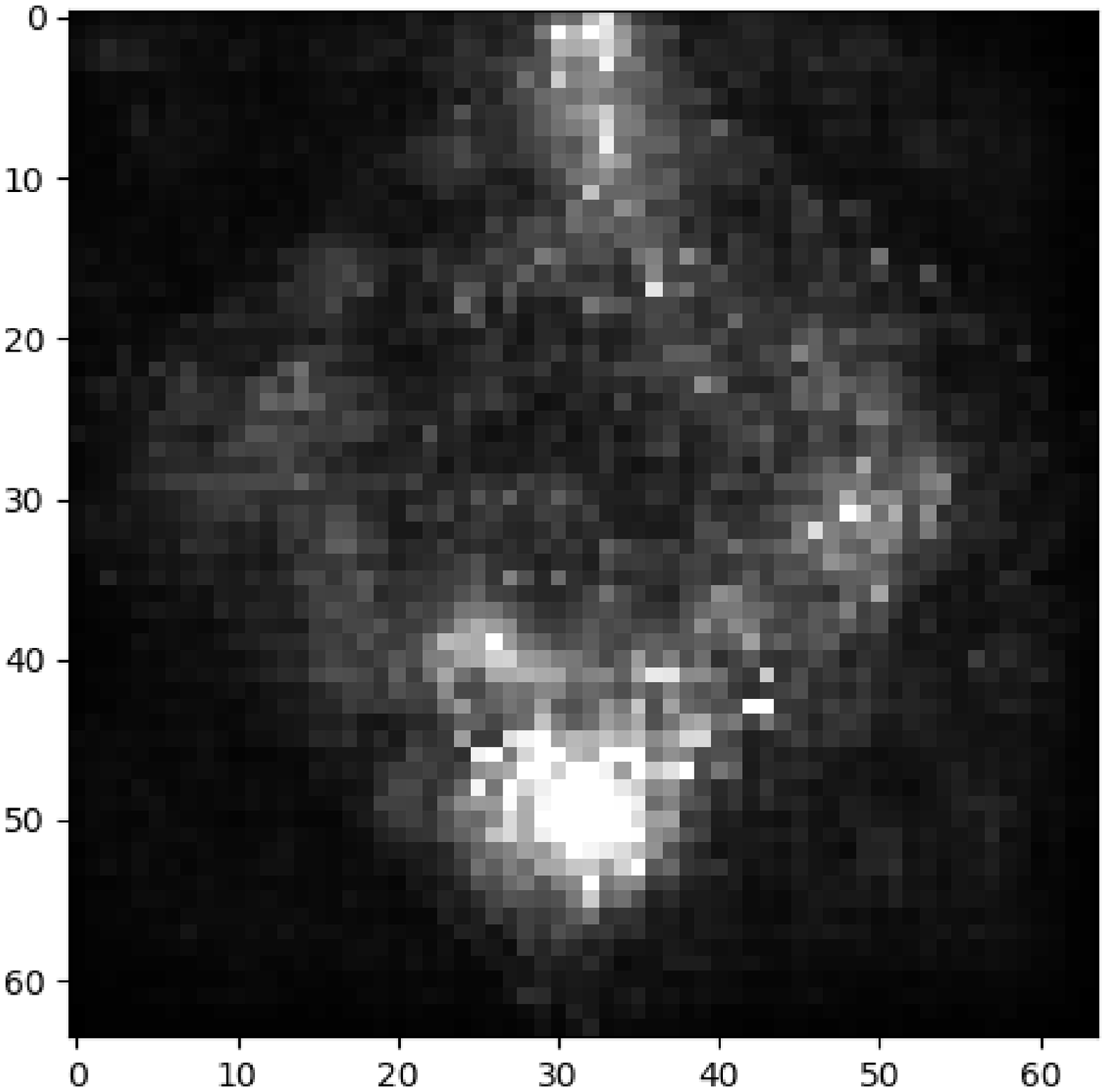} 
\label{fu2} 
\end{minipage}%
\begin{minipage}[t]{0.16\linewidth} 
\centering 
\includegraphics[width=1.3in]{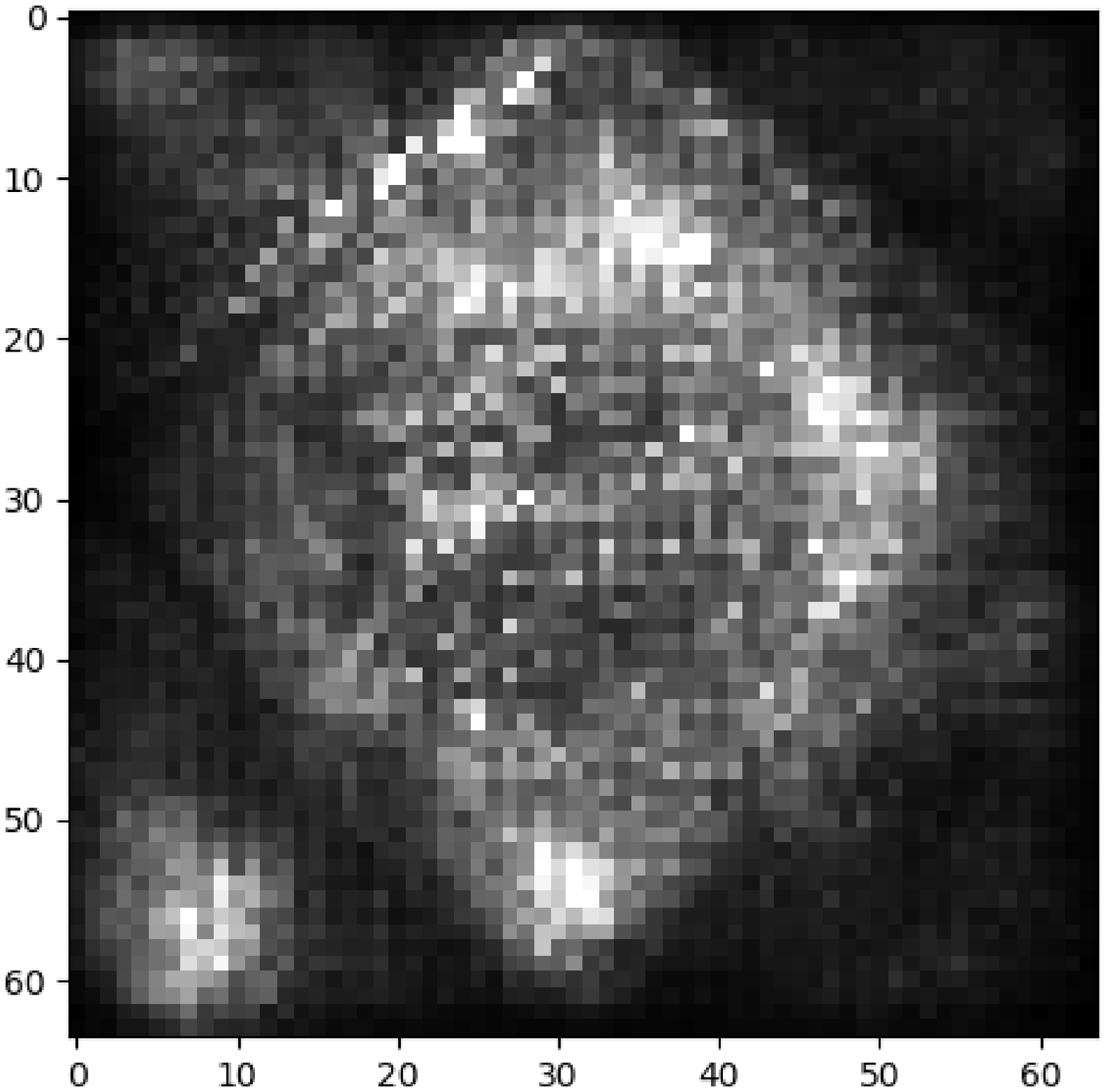} 
\label{fu4}
\end{minipage}%
\begin{minipage}[t]{0.16\linewidth} 
\centering 
\includegraphics[width=1.3in]{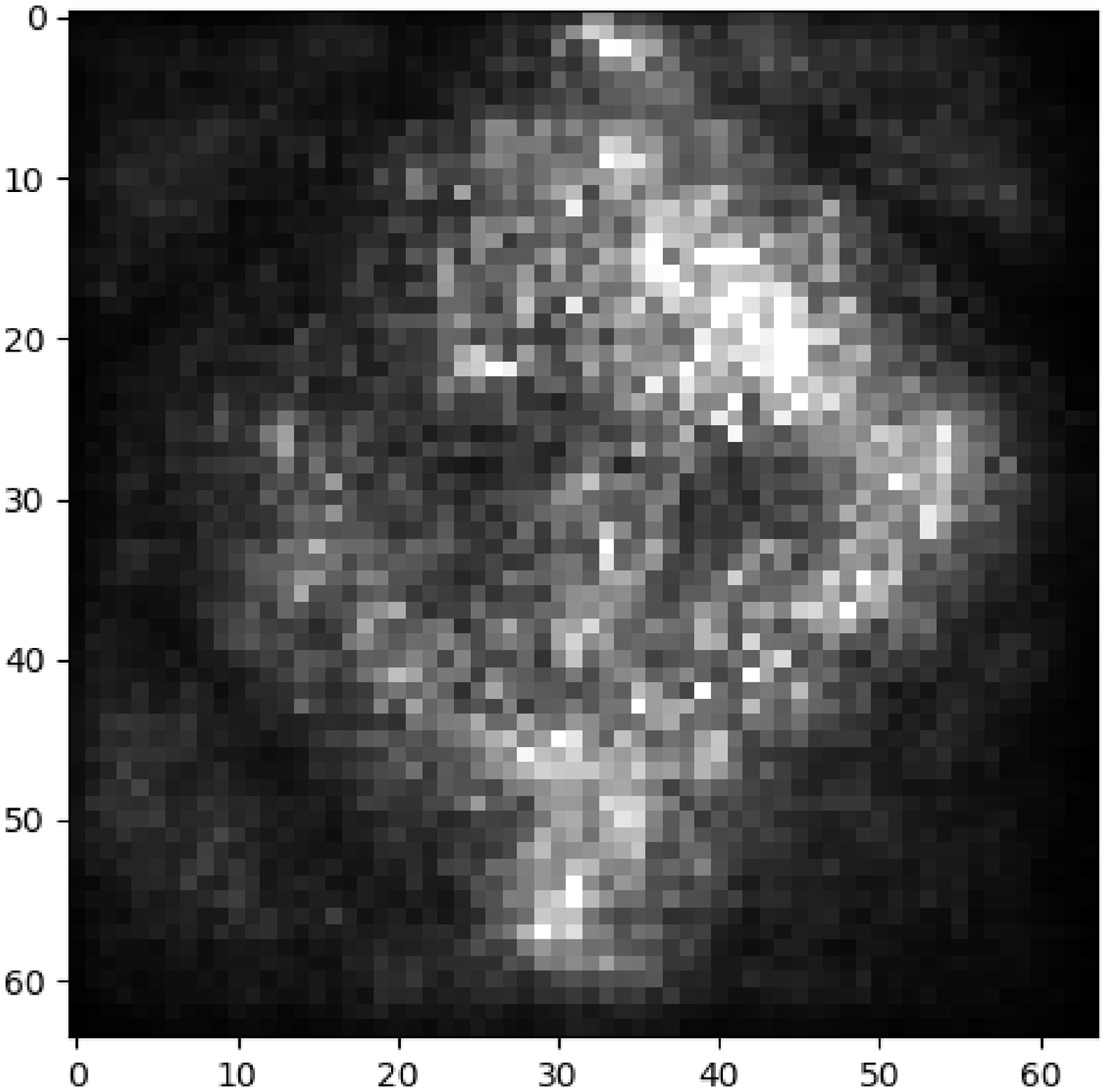} 
\label{fu3} 
\end{minipage}%
\begin{minipage}[t]{0.16\linewidth} 
\centering 
\includegraphics[width=1.3in]{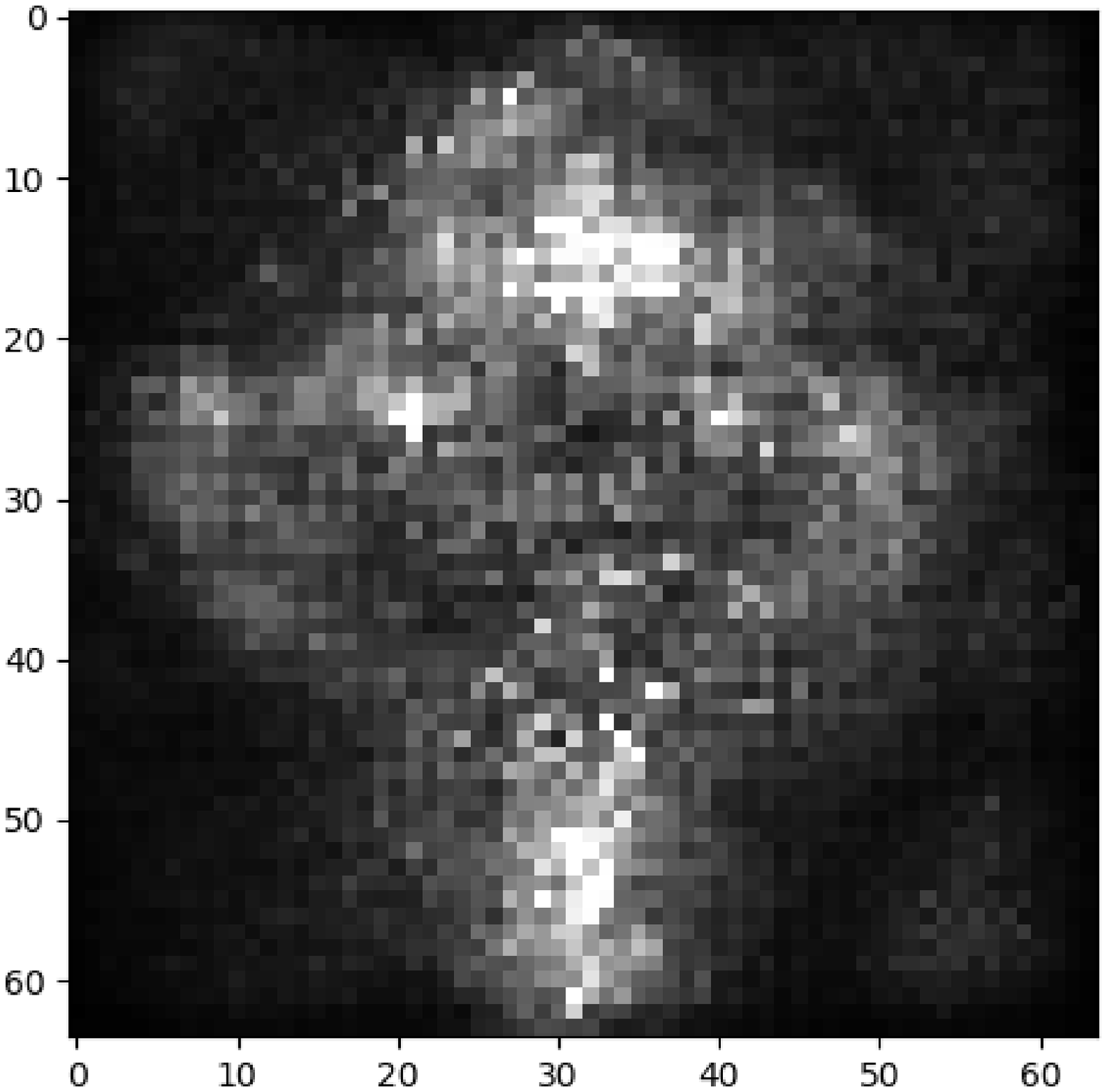} 
\label{fu5} 
\end{minipage}%

\caption{Pictures from left to right show the initial image, saliency maps obtained from Resnet34, ResNet34-MLP+, Resnet34-A+, Resnet34-MLPN, Resnet34-AN.}
\label{saliency_maps}
\end{figure*}

\begin{table*}[htb]
\small
\centering
\begin{tabular}{|l|l|l|l|l|l|l|l|l|l|l|}
\hline
\multicolumn{1}{|c|}{\multirow{2}{*}{Models}} &
  \multicolumn{1}{c|}{\multirow{2}{*}{Ori}} &
  \multicolumn{3}{c|}{White box   evaluation} &
  \multicolumn{6}{c|}{Black box   evaluation} \\ \cline{3-11} 
\multicolumn{1}{|c|}{} &
  \multicolumn{1}{c|}{} &
  FGSM &
  L-BFGS &
  PGD &
  Gaussian &
  Impluse &
  Glass Blur &
  Contrast &
  FGSM &
  L-BFGS \\ \hline
ResNet18       & 0.902 & 0.234 & 0.433 & 0.114 & 0.558 & 0.530  & \textbf{0.189} & 0.544 & 0.467 & 0.562 \\ \hline
ResNet18-MLP+ & 0.874 & 0.254 & 0.461 & 0.118 & 0.553 & 0.535 & 0.185          & 0.536 & 0.469 & 0.554 \\ \hline
ResNet18-A+    & 0.877 & 0.219 & 0.401 & 0.143 & 0.572 & 0.543 & 0.184          & 0.544 & 0.493 & 0.570  \\ \hline
ResNet18-MLPN  & 0.899 & 0.368 & 0.450  & 0.181 & 0.553 & 0.514 & 0.175          & 0.533 & 0.482 & 0.584 \\ \hline
ResNet18-AN &
  \textbf{0.905} &
  \textbf{0.393} &
  \textbf{0.489} &
  \textbf{0.203} &
  \textbf{0.587} &
  \textbf{0.557} &
  0.175 &
  \textbf{0.559} &
  \textbf{0.562} &
  \textbf{0.613} \\ \hline
\end{tabular}
\caption{Evaluation results for ResNet18 models in Cifar-10}
\label{resnet_cifar}
\end{table*}

\begin{table*}[htb]
\centering
\small
\begin{tabular}{|l|l|l|l|l|l|l|l|l|l|l|}
\hline
\multicolumn{1}{|c|}{\multirow{2}{*}{Models}} &
  \multicolumn{1}{c|}{\multirow{2}{*}{Ori}} &
  \multicolumn{3}{c|}{White box evaluation} &
  \multicolumn{6}{c|}{Black box   evaluation} \\ \cline{3-11} 
\multicolumn{1}{|c|}{} &
  \multicolumn{1}{c|}{} &
  FGSM &
  L-BFGS &
  PGD &
  Gaussian &
  Impluse &
  Glass Blur &
  Contrast &
  FGSM &
  L-BFGS \\ \hline
ResNet34       & 0.436 & 0.082 & 0.321          & 0.019 & 0.397 & 0.351 & 0.341 & 0.331 & 0.374          & 0.329 \\ \hline
ResNet34-MLP+ & 0.434 & 0.076 & 0.324          & 0.022 & 0.383 & 0.339 & 0.323 & 0.333 & 0.362          & 0.312 \\ \hline
ResNet34-A+    & 0.177 & 0.012 & 0.145          & 0.011 & 0.165 & 0.155 & 0.138 & 0.133 & 0.158          & 0.145 \\ \hline
ResNet34-MLPN  & 0.445 & 0.119 & \textbf{0.402} & 0.051 & 0.406 & 0.364 & 0.336 & 0.339 & \textbf{0.389} & 0.344 \\ \hline
ResNet34-AN &
  \textbf{0.448} &
  \textbf{0.121} &
  \textbf{0.402} &
  \textbf{0.055} &
  \textbf{0.412} &
  \textbf{0.375} &
  \textbf{0.352} &
  \textbf{0.346} &
  \textbf{0.389} &
  \textbf{0.350} \\ \hline
\end{tabular}
\caption{Evaluation results for ResNet34 models in tiny-ImageNet}
\label{resnet_tiny}
\end{table*}

\textbf{Robustness under white box attack}
The white box attack results are presented in Tables \ref{mlp_mnist} and \ref{cnn_mnist}. Adding standard normally distributed noises to either MLP or CNN can significantly improve the model's defensiveness for all of the FGSM, L-BFGS and PGD attacks. As a tradeoff, the accuracy on the original classification task is dropped to some extent. Surprisingly, by adding Gaussian noises optimized by our proposed method, we not only further improve the model's defensiveness at a decent margin, but also improve the classification accuracy in the original testing dataset. 

For MLP, it is interesting to notice that the Sigmoid activation function generally leads to better performance than the ReLu activation function. Compared to the base model MLP with the Sigmoid activation function, MLPN achieves a 20\%(0.336 vs 0.28) increase in accuracy under the FGSM attack, a 75\%(0.568 vs 0.324) increase in accuracy  under the L-BFGS attack, and a 33\%(0.275 vs 0.207)  increase in accuracy under the PGD attack. For CNN, it is interesting to observe that adding noises only to fully connected layers achieves better performance than adding noises to all  layers. CNN-MLPN achieves a 16\%(0.87 vs 0.774) increase in accuracy under the FGSM attack, a 11\%(0.685 vs 0.616) increase in accuracy under the L-BFGS attack, and a 15\%(0.752 vs 0.655) increase in accuracy under the PGD attack. Our method also increases the classification accuracy in the original dataset by 2.2\%(0.957 vs 0.936).

\textbf{Acceleration for training}
Our proposed method leads to a fast convergence speed in training an ANN ReLu activation functions. Fig.\ref{accelerationOnMLP} reports the training losses, validation losses, and accuracy in the testing dataset as a function of epochs. Compared to MLP and MLP+, MLPN leads to the fastest convergence speed and achieves a comparable classification accuracy. 

\textbf{Robustness under black box attack}
Robustness under black box attack is evaluated and shown in Tables \ref{mlp_mnist} and \ref{cnn_mnist}. To apply FGSM and L-BFGS, we use another MLP consisted of two hidden layers with 300 and 150 neurons and the Relu activation function at each layer to generate adversarial samples. Following ~\citet{Hendrycks2019noise}, we perform black box attacks by adding corruption noises to the images, including Gaussian, Impluse, Glass Blur, Contrast. The results show that standard normally distributed noises can improve the model's defensiveness against adversarial attacks but at a cost of a significant drop in accuracy in both the original testing dataset and the dataset corrupted by natural noises. On the other hand, our proposed noise optimization method achieves performance enhancement in all cases, i.e., accuracy in original dataset, and defensiveness against both adversarial attacks and natural noise corruptions.

Again, MLP with the Sigmoid activation function performs better than that with ReLu. Under the adversarial attacks, MLPN achieves a 13\%(0.465 vs 0.41) increase in accuracy for FGSM  and a 5.2\%(0.788 vs 0.749)  increase in accuracy for L-BFGS. Under  natural noise corruptions, MLPN achieves a 7.1\%(0.946 vs 0.883) increase in accuracy for Gaussian, a 21\%(0.944 vs 0.783) increase in accuracy for Impulse, a 4.0\%(0.92 vs 0.885) increase in accuracy  for Glass Blur, and a 4.8\%(0.71 vs 0.676) increase in accuracy for Contrast.  For CNN, adding noises only to fully connected layers(CNN-MLPN) also achieves better performance than adding noises to all the layers(CNN-AN) in most situations. Compares to the baseline, CNN-MLPN achieves a 4.4\%(0.957 vs 0.917) increase in accuracy under the FGSM attack and a 5.0\%(0.818 vs 0.779) increase in accuracy under the L-BFGS attack. Under natural  noise corruptions, CNN-MLPN achieves a 1.2\%(0.995 vs 0.983) increase  in accuracy  for Gaussian, a 1.3\%(0.984 vs 0.971) increase  in accuracy  for Impulse, a 4.8\%(0.788 vs 0.752) increase  in accuracy  for Glass Blur, and a 0.9\%(0.853 vs 0.845) increase in accuracy for Contrast. Our proposed method also improves  classification accuracy in the original testing dataset by 0.4\%(0.990 vs 0.986).

\subsection{Results in Cifar-10 dataset}
We adopt ResNet18 as the base model for Cifar-10 classification. To compare the influence of adding noises to fully connected neural layers and convolution neural layers, we replace the last fully connected neural layer of 10 hidden neurons with three fully connected neural layers which  consist of 256, 128 and 10  neurons at three layers, respectively. For convolution neural layers, we only add noises to the last convolution layer of each residual block. For generating adversarial samples in black box attack, we use the original ResNet18 with one fully connected neural layer to generate adversarial examples for both FGSM and L-BFGS. We randomly split the entire dataset into training, validation, and testing datasets in a ratio of 4:1:1.

The results are shown in Table \ref{resnet_cifar}. Notations ResNet18, ResNet18-MLP+, ResNet18-A+, ResNet18-MLPN, and ResNet18-AN are interpreted similarly as those in CNN described before. 
The results show that adding standard normally distributed noises does not improve robustness but deteriorate classification accuracy in the original testing dataset. However, adding noises optimized by our proposed method significantly improves the performance under both adversarial attacks and natural noise corruptions, as well as the classification accuracy in the original testing dataset. It is also worth noting that adding noises to both fully connected layer and convolution layer(ResNet18-AN) achieves the best performance in all cases. 

Compared to the baseline, ResNet18-AN achieves an average accuracy increase of 53\% under white box adversarial attacks ( 68\%(0.393 vs 0.234) for FGSM, 13\%(0.489 vs 0.433) for L-BFGS, 78\%(0.203 vs 0.114) for PGD), and an average accuracy increase of 9.7\% under black box adversarial attacks ( 20\%(0.562 vs 0.467) for FGSM, 9.1\%(0.613 vs 0.562) for L-BFGS). Under natural noise corruptions, although ResNet18-AN leads to an accuracy drop of 7.4\%(0.175 vs 0.189) for Glass Blur, it achieves significant performance improvement in defending the other three types of noises ( 5.2\%(0.587 vs 0.557) for Gaussian, 4.9\%(0.557 vs 0.53) for Impulse and 2.8\%(0.559 vs 0.544) for Contrast). ResNet18-AN also slightly improves  classification accuracy in the original testing dataset by 0.3\%(0.905 vs 0.902).

\subsection{Tiny-ImageNet}
We adopt ResNet34 as the base model for tiny-ImageNet classification. Tiny-ImageNet dataset~\cite{le2015tiny} is a subset of ImageNet which contains only 200 classes with 500 training images, 50 validation images and 50 test images in each class, and with the image size down-sampled to $64\times64\times3$ pixels. To compare the influence of adding noises to fully connected neural layers and convolution neural layers, we replace the last fully connected neural layer of 200 hidden neurons with four fully connected neural layers which consist of 1024, 512, 256 and 200 neurons at each layer, respectively. For convolution neural layers, we only add noises to the last convolution neural layer of each residual block. For generating adversarial samples in black box attack, we use the original ResNet34 with one fully connected neural layer to generate adversarial samples for both FGSM and L-BFGS. We randomly split the entire dataset into training, validation, and testing datasets in a ratio of 10:1:1.

The results are shown in Table \ref{resnet_tiny}. Notations ResNet34, ResNet34-MLP+, ResNet34-A+, ResNet34-MLPN, and ResNet34-AN are interpreted similarly as those described in CNN.
Similar to the observations in Cifar-10, adding standard normally distributed noises does not improve robustness but deteriorate accuracy, especially when noises are added to both the convolution neural layers and fully connected layers(ResNet34-A+), whereas our proposed method enhances performance in all cases with the best result achieved by ResNet34-AN. 

Compared to the baseline, ResNet34-AN achieves an average accuracy increase of 87\% under white box adversarial attacks ( 48\%(0.121 vs 0.082) for FGSM, 25\%(0.402 vs 0.321) for L-BFGS, 189\%(0.055 vs 0.019) for PGD), and an average accuracy increase of 5.2\% under black box adversarial attacks ( 4.0\%(0.389 vs 0.374) for FGSM, 6.4\%(0.35 vs 0.329) for L-BFGS). ResNet34-AN also has better defensiveness under all four types of natural noise corruptions, leading to an average accuracy increase of 4.6\% ( 3.8\%(0.412 vs 0.397) for Gaussian, 6.8\%(0.375 vs 0.351) for Impulse, 3.2\%(0.352 vs 0.341) for Glass Blur and 4.5\%(0.346 vs 0.331) for Contrast). ResNet34-AN improves  classification accuracy in the original testing dataset by 2.7\%(0.448 vs 0.436).

\subsection{Visualization on Saliency Map}
To help better understand why our noise optimization method improves robustness, we adopt SmoothGrad method ~\cite{Daniel2017SmoothGrad} to generate saliency maps for different models in tiny-ImageNet dataset. Gradient-based saliency map is typically used to represent 'saliency' at every location in the visual field, and it is adopted as a proxy for locating ``important'' pixels in the input image. The value on each pixel of the saliency map stands for the level of attention of the model.

For each sampled image, we add random noise $\mathcal{N}(0,\sigma^2)$ and generate the saliency map. We repeat the process $n$ times and average the saliency maps to obtain the final saliency map, which is computed by
\begin{equation}
\small
\begin{aligned}
    M^{(c)}(x) & = (\frac{\partial y^{c}}{\partial x})^{2},\\
    M^{(c)}_{smooth}(x) &= \frac{1}{n}\sum\limits^{n}_{1}M^{(c)}(x+\mathcal{N}(0,\sigma^2))
\end{aligned}
\end{equation}
where $y^{c}$ is the  $c$-th label's score (scalar output) given input $x$. We set $\sigma$ to $0.15$ and each image is reused  $n=25$ times. We show the 2d score by summarizing the pixels along three channels of the map. 

The results are shown in Fig. \ref{saliency_maps}. All of the images are sampled from testing dataset and classified correctly. Adding noises optimized by our proposed method makes the model focus more on the regions where targets are located and learn more important features. Taking the first picture as an example, we can see that ResNet34 concentrates all its attention on the face, whereas ResNet34-AN focuses on both the face and neck. Likewise for the other pictures, saliency maps of the ANNs with noises optimized by our method are more comprehensive and clear, which indicates that the ANNs capture more important features and thus lead to improved robustness in classification.

\section{Conclusion}
In this work, we propose a method to optimize the magnitudes of the noises added to ANN simultaneously in the process of training the synaptic weights at nearly no extra computation cost. 
Our method is applied to train both MLP and CNN with a ResNet backbone in MNIST, Cifar-10, and Tiny-ImageNet datasets. The proposed noise optimization method significantly improves the performance under both adversarial attacks and natural noise corruptions, as well as the classification accuracy in the original testing dataset. For training MLP, our method can also lead to a faster convergence speed in training. We use the saliency map to help better understand why our noise optimization method improves robustness.

\bibliography{citation}

\begin{thebibliography}{56}
\providecommand{\natexlab}[1]{#1}
\providecommand{\url}[1]{\texttt{#1}}
\expandafter\ifx\csname urlstyle\endcsname\relax
  \providecommand{\doi}[1]{doi: #1}\else
  \providecommand{\doi}{doi: \begingroup \urlstyle{rm}\Url}\fi

\bibitem[Asmussen \& Glynn(2007)Asmussen and Glynn]{asmussen2007stochastic}
Asmussen, S. and Glynn, P.~W.
\newblock \emph{Stochastic Simulation: Algorithms and Analysis}, volume~57.
\newblock Springer Science \& Business Media, 2007.

\bibitem[Athalye et~al.(2018)Athalye, Carlini, and Wagner]{CW_bpda}
Athalye, A., Carlini, N., and Wagner, D.
\newblock Obfuscated gradients give a false sense of security: Circumventing
  defenses to adversarial examples.
\newblock 2018.

\bibitem[Azulay \& Weiss(2019)Azulay and Weiss]{azulay2019why}
Azulay, A. and Weiss, Y.
\newblock Why do deep convolutional networks generalize so poorly to small
  image transformations?, 2019.
\newblock URL \url{https://openreview.net/forum?id=HJxYwiC5tm}.

\bibitem[Borji \& Lin(2019)Borji and Lin]{borji2019white}
Borji, A. and Lin, S.
\newblock White noise analysis of neural networks.
\newblock \emph{arXiv preprint arXiv:1912.12106}, 2019.

\bibitem[Brownlee(2019)]{brownlee2019train}
Brownlee, J.
\newblock Train neural networks with noise to reduce overfitting.
\newblock \emph{Machine Learning Mastery}, 2019.

\bibitem[Carlini \& Wagner(2017)Carlini and Wagner]{CW_ten}
Carlini, N. and Wagner, D.
\newblock Adversarial examples are not easily detected: Bypassing ten detection
  methods.
\newblock In \emph{Proceedings of the 10th ACM Workshop on Artificial
  Intelligence and Security}, pp.\  3--14, 2017.

\bibitem[Chan et~al.(2020)Chan, Tay, and Ong]{chan2020thinks}
Chan, A., Tay, Y., and Ong, Y.-S.
\newblock What it thinks is important is important: Robustness transfers
  through input gradients.
\newblock In \emph{Proceedings of the IEEE/CVF Conference on Computer Vision
  and Pattern Recognition}, pp.\  332--341, 2020.

\bibitem[Ciss{\'e} et~al.(2017)Ciss{\'e}, Bojanowski, Grave, Dauphin, and
  Usunier]{cisse2017parseval}
Ciss{\'e}, M., Bojanowski, P., Grave, E., Dauphin, Y., and Usunier, N.
\newblock Parseval networks: Improving robustness to adversarial examples.
\newblock 2017.

\bibitem[Dan~Hendrycks(2019{\natexlab{a}})]{Hendrycks2019noise}
Dan~Hendrycks, T.~D.
\newblock Benchmarking neural network robustness to common corruptions and
  perturbations.
\newblock In \emph{ICLR}, 2019{\natexlab{a}}.

\bibitem[Dan~Hendrycks(2019{\natexlab{b}})]{REN2020}
Dan~Hendrycks, T.~D.
\newblock Adversarial attacks and defenses in deep learning.
\newblock In \emph{Engineering}, volume~6, pp.\  346 -- 360,
  2019{\natexlab{b}}.
\newblock \doi{https://doi.org/10.1016/j.eng.2019.12.012}.

\bibitem[Daniel~Smilkov(2017)]{Daniel2017SmoothGrad}
Daniel~Smilkov, Nikhil~Thorat, B. K. F. V. M.~W.
\newblock Smoothgrad: removing noise by adding noise.
\newblock In \emph{arXiv preprint arXiv:1706.03825}, 2017.

\bibitem[Dhillon et~al.(2018)Dhillon, Azizzadenesheli, Lipton, Bernstein,
  Kossaifi, Khanna, and Anandkumar]{dhillon2018stochastic}
Dhillon, G.~S., Azizzadenesheli, K., Lipton, Z.~C., Bernstein, J., Kossaifi,
  J., Khanna, A., and Anandkumar, A.
\newblock Stochastic activation pruning for robust adversarial defense.
\newblock 2018.

\bibitem[Dong et~al.(2018)Dong, Liao, Pang, Su, Zhu, Hu, and
  Li]{dong2018boosting}
Dong, Y., Liao, F., Pang, T., Su, H., Zhu, J., Hu, X., and Li, J.
\newblock Boosting adversarial attacks with momentum.
\newblock In \emph{Proceedings of the IEEE conference on computer vision and
  pattern recognition}, pp.\  9185--9193, 2018.

\bibitem[Dong et~al.(2020)Dong, Fu, Yang, Pang, Su, Xiao, and
  Zhu]{dong2019benchmarking}
Dong, Y., Fu, Q.-A., Yang, X., Pang, T., Su, H., Xiao, Z., and Zhu, J.
\newblock Benchmarking adversarial robustness.
\newblock 2020.

\bibitem[Dziugaite et~al.(2016)Dziugaite, Ghahramani, and Roy]{jpeg}
Dziugaite, G.~K., Ghahramani, Z., and Roy, D.~M.
\newblock A study of the effect of {JPG} compression on adversarial images.
\newblock \emph{arXiv preprint arXiv:1608.00853}, 2016.

\bibitem[Gao et~al.(2020)Gao, Saha, Prasad, and Roychoudhury]{gao2020fuzz}
Gao, X., Saha, R.~K., Prasad, M.~R., and Roychoudhury, A.
\newblock Fuzz testing based data augmentation to improve robustness of deep
  neural networks.
\newblock In \emph{2020 IEEE/ACM 42nd International Conference on Software
  Engineering (ICSE)}, pp.\  1147--1158. IEEE, 2020.

\bibitem[Goodfellow et~al.(2014)Goodfellow, Shlens, and
  Szegedy]{goodfellow2014explaining}
Goodfellow, I.~J., Shlens, J., and Szegedy, C.
\newblock Explaining and harnessing adversarial examples.
\newblock \emph{arXiv preprint arXiv:1412.6572}, 2014.

\bibitem[Gulcehre et~al.(2016)Gulcehre, Moczulski, Denil, and
  Bengio]{gulcehre2016noisy}
Gulcehre, C., Moczulski, M., Denil, M., and Bengio, Y.
\newblock Noisy activation functions.
\newblock In \emph{International conference on machine learning}, pp.\
  3059--3068. PMLR, 2016.

\bibitem[Guo et~al.(2019)Guo, Gardner, You, Wilson, and
  Weinberger]{guo2019simple}
Guo, C., Gardner, J., You, Y., Wilson, A.~G., and Weinberger, K.
\newblock Simple black-box adversarial attacks.
\newblock In \emph{International Conference on Machine Learning}, pp.\
  2484--2493. PMLR, 2019.

\bibitem[Hadash et~al.(2018)Hadash, Kermany, Carmeli, Lavi, Kour, and
  Jacovi]{hadash2018estimate}
Hadash, G., Kermany, E., Carmeli, B., Lavi, O., Kour, G., and Jacovi, A.
\newblock Estimate and replace: A novel approach to integrating deep neural
  networks with existing applications.
\newblock \emph{arXiv preprint arXiv:1804.09028}, 2018.

\bibitem[Hang et~al.(2020)Hang, Han, Chen, and Li]{hang2020ensemble}
Hang, J., Han, K., Chen, H., and Li, Y.
\newblock Ensemble adversarial black-box attacks against deep learning systems.
\newblock volume 101, pp.\  107184. Elsevier, 2020.

\bibitem[He et~al.(2019)He, Yang, Li, Li, Chang, and Yu]{he2019non}
He, X., Yang, S., Li, G., Li, H., Chang, H., and Yu, Y.
\newblock Non-local context encoder: Robust biomedical image segmentation
  against adversarial attacks.
\newblock volume~33, pp.\  8417--8424, 2019.

\bibitem[Heaven(2019)]{heaven2019deep}
Heaven, D.
\newblock Why deep-learning ais are so easy to fool.
\newblock \emph{Nature}, 574\penalty0 (7777):\penalty0 163--166, 2019.

\bibitem[Heidergott \& Leahu(2010)Heidergott and Leahu]{heidergott2010weak}
Heidergott, B. and Leahu, H.
\newblock Weak differentiability of product measures.
\newblock \emph{Mathematics of Operations Research}, 35\penalty0 (1):\penalty0
  27--51, 2010.

\bibitem[Hendrycks et~al.(2019)Hendrycks, Mu, Cubuk, Zoph, Gilmer, and
  Lakshminarayanan]{hendrycks2019augmix}
Hendrycks, D., Mu, N., Cubuk, E.~D., Zoph, B., Gilmer, J., and
  Lakshminarayanan, B.
\newblock Augmix: A simple data processing method to improve robustness and
  uncertainty.
\newblock \emph{arXiv preprint arXiv:1912.02781}, 2019.

\bibitem[Ho \& Cao(1991)Ho and Cao]{ho1991discrete}
Ho, Y.-C. and Cao, X.-R.
\newblock \emph{Discrete Event Dynamic Systems and Perturbation Analysis}.
\newblock Kluwer Academic Publishers, Boston, MA, 1991.

\bibitem[Hong(2009)]{hong2009estimating}
Hong, L.~J.
\newblock Estimating quantile sensitivities.
\newblock \emph{Operations Research}, 57\penalty0 (1):\penalty0 118--130, 2009.

\bibitem[Krizhevsky et~al.(2012)Krizhevsky, Sutskever, and
  Hinton]{krizhevsky2012imagenet}
Krizhevsky, A., Sutskever, I., and Hinton, G.~E.
\newblock Imagenet classification with deep convolutional neural networks.
\newblock \emph{Advances in neural information processing systems},
  25:\penalty0 1097--1105, 2012.

\bibitem[Le \& Yang(2015)Le and Yang]{le2015tiny}
Le, Y. and Yang, X.
\newblock Tiny imagenet visual recognition challenge.
\newblock \emph{CS 231N}, 7:\penalty0 7, 2015.

\bibitem[Liao et~al.()Liao, Liang, Dong, Pang, Hu, and Zhu]{liao2018defense}
Liao, F., Liang, M., Dong, Y., Pang, T., Hu, X., and Zhu, J.
\newblock Defense against adversarial attacks using high-level representation
  guided denoiser.

\bibitem[Ling et~al.(2019)Ling, Ji, Zou, Wang, Wu, Li, and
  Wang]{ling2019deepsec}
Ling, X., Ji, S., Zou, J., Wang, J., Wu, C., Li, B., and Wang, T.
\newblock Deepsec: A uniform platform for security analysis of deep learning
  model.
\newblock In \emph{2019 IEEE Symposium on Security and Privacy (SP)}, pp.\
  673--690. IEEE, 2019.

\bibitem[Liu et~al.(2018)Liu, Cheng, Zhang, and Hsieh]{liu2018towards}
Liu, X., Cheng, M., Zhang, H., and Hsieh, C.-J.
\newblock Towards robust neural networks via random self-ensemble.
\newblock pp.\  369--385, 2018.

\bibitem[Madry et~al.(2017)Madry, Makelov, Schmidt, Tsipras, and
  Vladu]{madry2017towards}
Madry, A., Makelov, A., Schmidt, L., Tsipras, D., and Vladu, A.
\newblock Towards deep learning models resistant to adversarial attacks.
\newblock \emph{arXiv preprint arXiv:1706.06083}, 2017.

\bibitem[Madry et~al.(2018)Madry, Makelov, Schmidt, Tsipras, and Vladu]{PGD}
Madry, A., Makelov, A., Schmidt, L., Tsipras, D., and Vladu, A.
\newblock Towards deep learning models resistant to adversarial attacks.
\newblock 2018.

\bibitem[Mohamed et~al.(2020)Mohamed, Rosca, Figurnov, and
  Mnih]{mohamed2019monte}
Mohamed, S., Rosca, M., Figurnov, M., and Mnih, A.
\newblock Monte {C}arlo gradient estimation in machine learning.
\newblock \emph{Journal of Machine Learning Research}, 21\penalty0
  (132):\penalty0 1--62, 2020.

\bibitem[Nazemi \& Fieguth(2019)Nazemi and Fieguth]{nazemi2019potential}
Nazemi, A. and Fieguth, P.
\newblock Potential adversarial samples for white-box attacks.
\newblock 2019.

\bibitem[Neelakantan et~al.(2015)Neelakantan, Vilnis, Le, Sutskever, Kaiser,
  Kurach, and Martens]{neelakantan2015adding}
Neelakantan, A., Vilnis, L., Le, Q.~V., Sutskever, I., Kaiser, L., Kurach, K.,
  and Martens, J.
\newblock Adding gradient noise improves learning for very deep networks.
\newblock \emph{arXiv preprint arXiv:1511.06807}, 2015.

\bibitem[Papernot et~al.(2016)Papernot, McDaniel, Wu, Jha, and
  Swami]{papernot2016distillation}
Papernot, N., McDaniel, P., Wu, X., Jha, S., and Swami, A.
\newblock Distillation as a defense to adversarial perturbations against deep
  neural networks.
\newblock In \emph{IEEE Symposium on Security and Privacy}, pp.\  582--597.
  IEEE, 2016.

\bibitem[Papernot et~al.(2017)Papernot, McDaniel, Goodfellow, Jha, Celik, and
  Swami]{masking}
Papernot, N., McDaniel, P., Goodfellow, I., Jha, S., Celik, Z.~B., and Swami,
  A.
\newblock Practical black-box attacks against machine learning.
\newblock In \emph{ASIA Computer and Communications Security}, pp.\  506--519,
  2017.

\bibitem[Parkhi et~al.(2015)Parkhi, Vedaldi, and Zisserman]{parkhi2015deep}
Parkhi, O.~M., Vedaldi, A., and Zisserman, A.
\newblock Deep face recognition.
\newblock In \emph{British Machine Vision Conference}, 2015.

\bibitem[Peng et~al.(2018)Peng, Fu, Hu, and Heidergott]{peng2015discontinuity}
Peng, Y., Fu, M.~C., Hu, J.-Q., and Heidergott, B.
\newblock A new unbiased stochastic derivative estimator for discontinuous
  sample performances with structural parameters.
\newblock \emph{Operations Research}, 66\penalty0 (2):\penalty0 487--499, 2018.

\bibitem[Petrov \& Hospedales(2019)Petrov and Hospedales]{petrov2019measuring}
Petrov, D. and Hospedales, T.~M.
\newblock Measuring the transferability of adversarial examples.
\newblock \emph{arXiv preprint arXiv:1907.06291}, 2019.

\bibitem[Prabhu \& Whaley()Prabhu and Whaley]{prabhu2018grey}
Prabhu, V.~U. and Whaley, J.
\newblock On grey-box adversarial attacks and transfer learning.
\newblock \emph{online: https://unify. id/wpcontent/uploads/2018/03/greybox
  attack. pdf}.

\bibitem[Ross \& Doshi-Velez(2017)Ross and Doshi-Velez]{ross2017improving}
Ross, A.~S. and Doshi-Velez, F.
\newblock Improving the adversarial robustness and interpretability of deep
  neural networks by regularizing their input gradients.
\newblock 2017.

\bibitem[Rubinstein \& Shapiro(1993)Rubinstein and
  Shapiro]{rubinstein1993discrete}
Rubinstein, R.~Y. and Shapiro, A.
\newblock \emph{Discrete Event Systems: Sensitivity Analysis and Stochastic
  Optimization by the Score Function Method}.
\newblock Wiley, New York, 1993.

\bibitem[Szegedy et~al.(2014)Szegedy, Zaremba, Sutskever, Bruna, Erhan,
  Goodfellow, and Fergus]{szegedy2013intriguing}
Szegedy, C., Zaremba, W., Sutskever, I., Bruna, J., Erhan, D., Goodfellow, I.,
  and Fergus, R.
\newblock Intriguing properties of neural networks.
\newblock 2014.

\bibitem[Taghanaki et~al.(2019)Taghanaki, Abhishek, Azizi, and
  Hamarneh]{taghanaki2019kernelized}
Taghanaki, S.~A., Abhishek, K., Azizi, S., and Hamarneh, G.
\newblock A kernelized manifold mapping to diminish the effect of adversarial
  perturbations.
\newblock pp.\  11340--11349, 2019.

\bibitem[Tram{\`e}r et~al.(2018)Tram{\`e}r, Kurakin, Papernot, Goodfellow,
  Boneh, and McDaniel]{tramer2017ensemble}
Tram{\`e}r, F., Kurakin, A., Papernot, N., Goodfellow, I., Boneh, D., and
  McDaniel, P.
\newblock Ensemble adversarial training: Attacks and defenses.
\newblock 2018.

\bibitem[Tramer et~al.(2020)Tramer, Carlini, Brendel, and
  Madry]{tramer2020adaptive}
Tramer, F., Carlini, N., Brendel, W., and Madry, A.
\newblock On adaptive attacks to adversarial example defenses.
\newblock 2020.

\bibitem[Tramèr et~al.(2017)Tramèr, Papernot, Goodfellow, Boneh, and
  McDaniel]{Florian207}
Tramèr, F., Papernot, N., Goodfellow, I., Boneh, D., and McDaniel, P.
\newblock The space of transferable adversarial examples.
\newblock \emph{arXiv}, 2017.
\newblock URL \url{https://arxiv.org/abs/1704.03453}.

\bibitem[Vasiljevic et~al.(2016)Vasiljevic, Chakrabarti, and
  Shakhnarovich]{vasiljevic2016examining}
Vasiljevic, I., Chakrabarti, A., and Shakhnarovich, G.
\newblock Examining the impact of blur on recognition by convolutional
  networks.
\newblock \emph{arXiv preprint arXiv:1611.05760}, 2016.

\bibitem[Xiang et~al.(2020)Xiang, Xu, Li, Ma, Xuan, and Liu]{xiang2020side}
Xiang, Y., Xu, Y., Li, Y., Ma, W., Xuan, Q., and Liu, Y.
\newblock Side-channel gray-box attack for dnns.
\newblock \emph{IEEE Transactions on Circuits and Systems II: Express Briefs},
  2020.

\bibitem[Xiao et~al.(2019)Xiao, Peng, Hong, Ke, and Yang]{xiao2019training}
Xiao, L., Peng, Y., Hong, J., Ke, Z., and Yang, S.
\newblock Training artificial neural networks by generalized likelihood ratio
  method: Exploring brain-like learning to improve robustness.
\newblock \emph{arXiv preprint arXiv:1902.00358}, 2019.

\bibitem[Xu et~al.(2017)Xu, Evans, and Qi]{xu2017feature}
Xu, W., Evans, D., and Qi, Y.
\newblock Feature squeezing: Detecting adversarial examples in deep neural
  networks.
\newblock In \emph{Network and Distributed System Security Symposium}, 2017.

\bibitem[You et~al.(2019)You, Ye, Li, Xu, and Wang]{you2019adversarial}
You, Z., Ye, J., Li, K., Xu, Z., and Wang, P.
\newblock Adversarial noise layer: Regularize neural network by adding noise.
\newblock In \emph{2019 IEEE International Conference on Image Processing
  (ICIP)}, pp.\  909--913. IEEE, 2019.

\bibitem[Zheng et~al.(2016)Zheng, Song, Leung, and
  Goodfellow]{zheng2016improving}
Zheng, S., Song, Y., Leung, T., and Goodfellow, I.
\newblock Improving the robustness of deep neural networks via stability
  training.
\newblock In \emph{Proceedings of the ieee conference on computer vision and
  pattern recognition}, pp.\  4480--4488, 2016.

\end{thebibliography}
\small
\bibliographystyle{icml2021}


\end{document}